\documentclass{article}
\usepackage{ijcai25}

\usepackage{times}
\usepackage{soul}
\usepackage{url}
\usepackage[hidelinks]{hyperref}
\usepackage[utf8]{inputenc}
\usepackage[small]{caption}
\usepackage{graphicx}
\usepackage{amsmath}
\usepackage{amsthm}
\usepackage{booktabs}
\usepackage{algorithm}
\usepackage{algorithmic}
\usepackage[switch]{lineno}

\usepackage{amsmath,amssymb,amsfonts}
\usepackage{float}
\usepackage{subfig}
\usepackage{bm}
\usepackage{url}
\usepackage{booktabs}
\usepackage{algorithm}
\usepackage{algorithmic}
\usepackage{comment}
\usepackage{multirow}

\title{Backdoor Attack on Vertical Federated Graph Neural Network Learning}

\author{
    Jirui Yang$^{1}$ \and
    Peng Chen$^{2}$ \and
    Zhihui Lu$^{1}$ \and
    Jianping Zeng$^{1}$ \and
    Qiang Duan$^{3}$ \and
    Xin Du$^{4}$ \and
    Ruijun Deng$^{1}$ \\
    \affiliations
    $^1$Fudan University, China\\
    $^2$Nanjing University of Information Science and Technology, China\\
    $^3$Pennsylvania State University, USA\\
    $^4$Zhejiang University, China\\
    \emails
    \{yangjr23, lzh, zjp, rjdeng22\}@m.fudan.edu.cn, 003913@nuist.edu.cn, qduan@psu.edu, xindu@zju.edu.cn
}

\begin{document}

\maketitle

\begin{abstract}

Federated Graph Neural Network (FedGNN) integrate federated learning (FL) with graph neural networks (GNNs) to enable privacy-preserving training on distributed graph data. Vertical Federated Graph Neural Network (VFGNN), a key branch of FedGNN, handles scenarios where data features and labels are distributed among participants. Despite the robust privacy-preserving design of VFGNN, we have found that it still faces the risk of backdoor attacks, even in situations where labels are inaccessible. 
This paper proposes BVG, a novel backdoor attack method that leverages multi-hop triggers and backdoor retention, requiring only four target-class nodes to execute effective attacks.
Experimental results demonstrate that BVG achieves nearly 100\% attack success rates across three commonly used datasets and three GNN models, with minimal impact on the main task accuracy. 
We also evaluated various defense methods, and the BVG method maintained high attack effectiveness even under existing defenses. 
This finding highlights the need for advanced defense mechanisms to counter sophisticated backdoor attacks in practical VFGNN applications.

\end{abstract}

%

\section{Introduction}

Graph Neural Networks (GNNs), with their powerful capability to process graph-structured data, have demonstrated significant value in cross-domain applications such as bioinformatics, chemical analysis, medical diagnosis, and financial risk management \cite{wu2020comprehensive}. Specifically, in financial fraud detection scenarios, the use of transaction relationship graphs to identify potential fraudulent activities highlights the unique advantages of GNNs \cite{cheng2023anti}. However, in real-world settings, graph data is often distributed across different stakeholders, and due to data privacy regulations and the need to protect commercial secrets, traditional centralized training paradigms face severe challenges \cite{FedGNN-survey-TNNLS24}.

\begin{figure}
    \centering
    \includegraphics[width=1\linewidth]{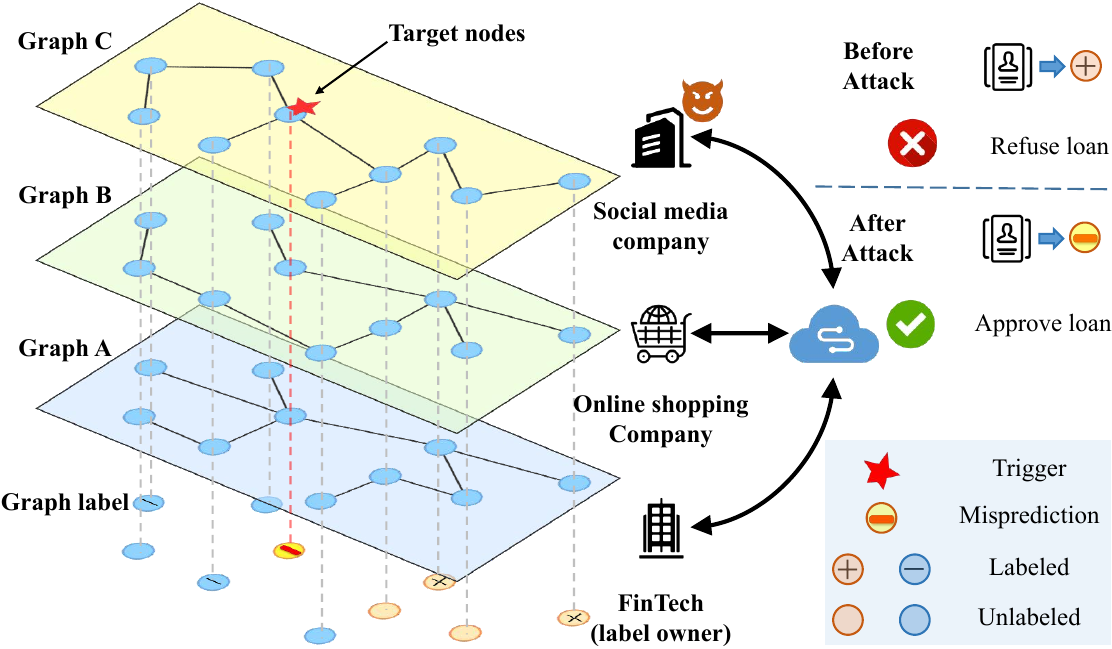}
    \caption{An example of VFGNN backdoor attack}
    \label{fig:exa}
\end{figure}

Federated Graph Neural Networks (FedGNN), as a combination of Federated Learning (FL) and GNNs, provide an innovative solution to this challenge \cite{he2021fedgraphnn}. Among them, Vertical Federated Graph Neural Networks (VFGNN) demonstrate unique value in cross-organizational collaboration scenarios due to their ability to handle heterogeneous distributions of features and labels \cite{mai2023vertical}. For instance, as shown in Figure~\ref{fig:exa}, in a credit evaluation system, financial institutions hold user transaction records and credit labels, social media platforms possess social graphs, and e-commerce platforms provide consumer behavior features. This collaborative multi-party data setting is a typical application scenario for VFGNN.

Although VFGNN effectively protects privacy in distributed scenarios, it also faces significant security challenges, with backdoor attacks being a major potential threat. Attackers can implant concealed trigger patterns to induce specific inputs to produce predefined malicious outputs while maintaining the model's normal predictive performance, posing severe risks to the practical application of VFGNN. For example, in the credit evaluation scenario illustrated in Figure~\ref{fig:exa}, malicious participants could use backdoor attacks to enable high-risk users to obtain credit approvals, leading to serious financial security issues. However, research on backdoor attacks targeting VFGNN remains nearly nonexistent. This is primarily due to two major challenges faced by traditional backdoor attacks in VFGNN scenarios: first, the label isolation mechanism among participants restricts traditional attack paths based on label manipulation; second, the complex topology of graph data requires backdoor triggers to not only maintain structural consistency but also satisfy concealment requirements.

To address the above challenges, this paper proposes a novel graph-structured backdoor attack method based on multi-hop adjacency relationships, termed BVG. This method innovatively designs a backdoor injection mechanism tailored for VFGNNs. First, it employs a multi-hop adjacency trigger generation algorithm to covertly propagate target-class features within the graph structure. Second, a backdoor retention strategy is adopted to enhance the attack's stability. Experimental results demonstrate that BVG requires knowledge of only four target-class nodes to achieve an attack success rate of over 99\%, with minimal impact on the performance of the main task. The main contributions of this paper are as follows:

\begin{itemize}
\item \textbf{Reveal backdoor attack risks in VFGNNs}: A systematic analysis of security threats within the VFGNN framework highlights unique risk characteristics distinct from traditional scenarios.
\item \textbf{Design multi-hop graph structural triggers}: A bi-level optimization strategy generates backdoor triggers for graph propagation, enabling flexible and covert backdoor injection without disrupting the graph structure.
\item \textbf{Develop a backdoor retention strategy}: This effectively addresses instability issues in backdoor injection during federated training, improving attack persistence.
\item \textbf{Validate effectiveness through experiments}: Extensive experiments on widely-used benchmark datasets demonstrate that BVG is efficient and stable, maintaining high attack success rates even under various backdoor defense mechanisms.
\end{itemize}

\section{Related Work} \label{related-work}


In the fields of VFL and GNN, researchers are investigating backdoor attacks specifically tailored to each domain.


\subsection{Backdoor attacks on VFL}


Federated Learning is especially vulnerable to backdoor attacks due to its distributed nature \cite{zhang2024a3fl}. However, the vertically split model of VFL restricts the attacker's access to the training labels, let alone modify them. 

There are two notable directions in attempting to solve this issue. The first is to conduct label inference attacks \cite{277244} to get some labels in advance. 
For example, VILLAIN \cite{bai2023villain} leverages label inference to pinpoint samples of the target label and then poisons these samples to inject the backdoor. BadVFL \cite{naseri2023badvfl} and LR-BA \cite{DBLP:journals/compsec/GuB23} assume attackers can acquire labels of a certain number of samples from each category with label inference. Another direction is to make the knowledge of a small number of labeled samples from the target class a prerequisite \cite{chen2024universal,chen2023practical,he2023backdoor}. These methods try to use as few as possible (e.g., 500 for \cite{chen2024universal} and 0.1\% for \cite{chen2023practical}) the labeled target-class samples to establish links between backdoor triggers and target labels. Normally, in VFL, the backdoor attack can only be conducted in a clean-label manner \cite{zhao2020clean}. Therefore, works like TPGD \cite{DBLP:conf/nips/LiuXK022} assuming the label modification capability of the active party are impractical.

\subsection{Backdoor attacks on GNN}

The unique challenge to graph-oriented backdoor attacks lies in the inherently unstructured and discrete nature of graph data \cite{xi2021graph}. For structured and continuous data like images, attackers can directly stamp a trigger pattern (e.g., a black square on the bottom right of the image) $s$ onto a benign image $x$ to achieve a poisoned sample $x_p = x+s$ \cite{li2022backdoor}. However, the backdoor attacks against GNN involve design triggers within a large spectrum, including topological structures and descriptive (nodes and edges) features \cite{xi2021graph}. The effectiveness of backdoor attacks largely depends on designing triggers tailored to the specific attributes of the target task. These triggers mainly fall into three categories: malicious subgraph structures, graph structure perturbations, and node attribute manipulation.


Malicious subgraph structures involve adding malicious subgraphs to nodes \cite{dai2023unnoticeable}, edge endpoints \cite{zheng2023link}, or specific positions in the graph \cite{GNN-backdoor-ccfc21,xi2021graph}, causing nodes, edges, or graphs to be misclassified. Graph structure perturbation involves injecting special node connections into the training set adversarially, allowing backdoor attacks to be executed by merely modifying the graph's topology during the attack \cite{yang2022transferable}. Node attribute manipulation involves carefully designing special node features and adding them to the target class nodes in the training set, followed by adaptive optimization of the graph structure to train a GNN model with backdoors \cite{xing2023clean,chen2023feature}. Although these methods are viable in centralized GNN scenarios, they all rely on full access to the training set, which is nearly impossible in VFGNN.

The aforementioned backdoor techniques, whether tailored for GNN or VFL, are not applicable to VFGNN due to unique threat models. Existing methods such as CBA, DBA, and Bkd-FedGNN focus on HFGNN and do not address the needs of VFGNN \cite{xu2022more,liu2023bkd}. VFL backdoor attack methods lack consideration for the data structure of graph data, while GNN backdoor attacks face difficulties in handling vertically partitioned data. 
To address these issues, this paper proposes the BVG algorithm for VFGNN, leveraging multi-hop triggers and a backdoor retention strategy to achieve efficient backdoor attacks.



\section{Proposed Approach} \label{model}

\begin{figure*}[!tbp]
\centerline{\includegraphics[width=1\linewidth]{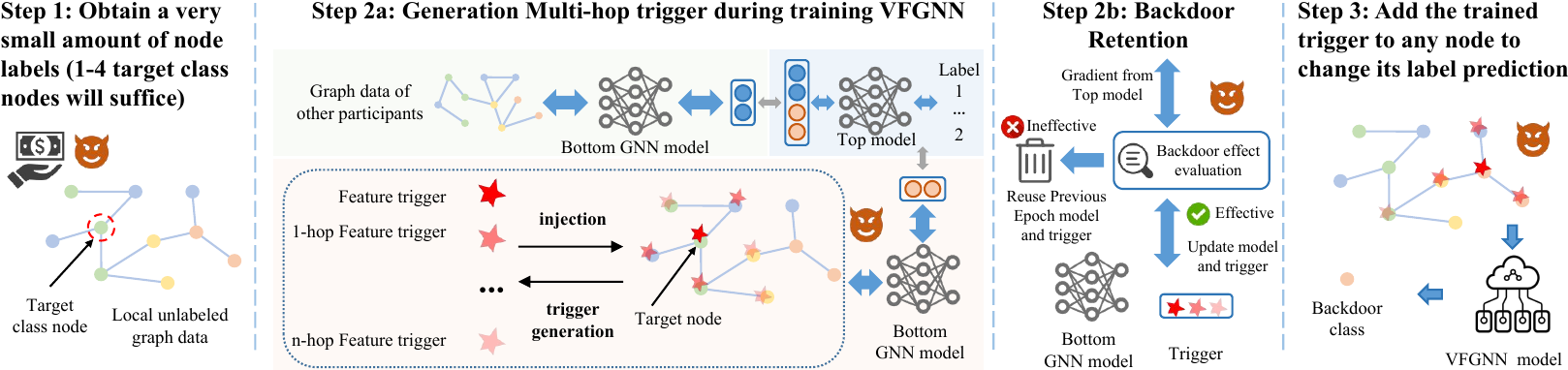}}
\caption{A sketch of our proposed Backdoor Attack for VFGNN.}
\label{fig:overview}
\end{figure*}

In this section, we formalize the backdoor attack problem in VFGNN and provide a detailed threat model. Next, we present our attack method, with a sketch of the entire approach shown in Figure \ref{fig:overview}.

\subsection{Problem Formulation}

Graph data is represented as $G = (V, E, X)$, where $V = \{v_1, \ldots, v_n\}$ is a set of N nodes, $E \subseteq V \times V$ is a set of edges, and $X = \{x_1, \ldots, x_N\}$ is a set of node attributes, with $x_i$ being the attribute of node $v_i$. The adjacency matrix of graph G is denoted as $A \in \mathbb{R}^{N \times N}$, where $A_{ij} = 1$ if nodes $v_i$ and $v_j$ are connected; otherwise, $A_{ij} = 0$.

We focus on the node classification problem, which is common in real-world applications like social networks. In inductive node classification tasks, only a subset of nodes \(V_L\) in the training graph have labels $Y_L = \{y_1, \ldots, y_{N_L}\}$. The test nodes \(V_T\) are disjoint from the training nodes.

In VFGNN, $K$ ($K>=2$) participants collaboratively train a model using their private data. Each participant $k$ has its own local graph $G^{k} = (V, E^{k}, X^{k})$, which comprises the entire node set $V$, a subset of edges $E^{k}$, and a subset of feature data $X^{k}$ \cite{chen2020vertically}. From a global perspective, all participants collectively own a global graph $G^{gl} = (V, E^{gl}, X^{gl})$, where $E^{gl} = E^{1} \cup \ldots \cup E^{K}$, and for any node attribute in $X$, $x_i^{gl} = x_i^1 || \ldots || x_i^K$, where $||$ is the concatenation operator. One of the $K$ participants is an active party that knows the corresponding labels of the labeled node set $V_L \subseteq V$, and all other participants are negative parties that have no access to the label information.   

Each participant employs a local GNN model $f_k$ parameterized by $\theta_k$ to compute the local output $H^k_i = f_{k}(\mathcal{G}_{i}^k;\theta_k)$, where $\mathcal{G}_{i}^k$ denotes the computation graph of node $v_i$. In addition to training its own local model, the active party also trains a top model $G$ parameterized by $\theta_{top}$, which aggregates the outputs from all local models to minimize the loss function $\mathcal{L}$. Therefore, the VFGNN model training can be formulated as
\begin{equation}\label{eq1}
\arg\min_{\Theta}\sum_{v_l\in V_L}\mathcal{L}(G(H^1_l, \cdots, H^K_l), y_l)
\end{equation}
where $\Theta = \{\theta_1, \cdots, \theta_K; \theta_{top}\}$ represents the parameters of the overall VFGNN model.

As an adversary in a backdoor attack, the main objectives are threefold: first, to ensure that the federated learning task can be completed successfully; second, to establish a mapping between the trigger and the backdoor target class; and third, to ensure that the backdoor injection remains undetected by the active party. These three objectives must be achieved simultaneously during the VFGNN training process. Specifically, the attacker needs to inject the backdoor cleverly, ensuring that the training process is not disrupted and that no suspicious behavior is exposed. After the VFGNN model is deployed, the attacker can introduce a local trigger to deliberately misclassify the corresponding sample as the target class, thereby completing the backdoor attack.
The backdoor attack objective in VFGNN is
\begin{equation}\label{eq2}
\begin{aligned}
    \min _{\Theta}  \underbrace{ \sum_{v_l\in V_L} \mathcal{L} \left(F(\mathcal{G}_{l}; \Theta), y_l\right)}_{\text{Main Task}}
    &+ \underbrace{\sum_{v_i\in V} \mathcal{L} \left(F(a(\mathcal{G}_{i}, \delta); \Theta), \tau\right)}_{\text{Backdoor Task}} \\
    &+ \underbrace{\sum_{v_i\in V}||\tilde{H}_i-H_i||^2}_{\text{Undetected Task}} 
\end{aligned}
\end{equation}
where $F$ refers to the VFGNN model, which encompasses both the top model and the bottom models of all participating parties. $\delta$ represents the backdoor trigger, a(·) denotes the operation of trigger attachment, and $\tau$ is the backdoor target class. 
$\tilde{H}$ denotes the output of the bottom GNN model after the trigger injection, while $H$ refers to the output without trigger injection.


\subsection{Threat model}
We assume the adversary is a passive party, not the active party, because the active party can modify labels and thus easily perform backdoor attacks. All other participants are considered trustworthy.

\textbf{Adversary's capacity.} 
The adversary adheres to the VFL protocol, transmitting local features and receiving gradients without manipulating other participants' information.

\textbf{Adversary's objective.}
In VFGNN multi-classification tasks,  the adversary's objective is to inject a backdoor into the model during the training phase without being detected. When the poisoned model is deployed for applications, it will misclassify any local data with the backdoor trigger as the designated target class but maintains classification capacity for all other clean data.  

\textbf{Adversary's knowledge.} 
In backdoor attacks, the adversary requires only a very few training samples labeled target class. Although this requirement deviates from the original VFGNN setting, it is feasible in practical VFGNN applications because it is possible for an adversary to acquire a small number of target class samples through various means, such as direct purchasing \cite{277244,DBLP:journals/compsec/GuB23}. Our experiment results indicate that acquiring only one to four target class nodes is sufficient for a successful backdoor attack. Apart from these limited target samples, the adversary has no information about the models and data of other parties.


\subsection{Multi-hop Trigger Generation}



Due to the special nature of the VFGNN architecture, all participants are aware of all nodes in the dataset, but possess different attributes of the nodes. Therefore, the adversary cannot generate a subgraph structure as a trigger like common GNN backdoor attacks \cite{dai2023unnoticeable,wang2024explanatory}. Here, the trigger is added to the node attributes. To prepare a trigger, we propose a multi-hop trigger generation method without introducing new nodes. The trigger generation and training are conducted synchronously with the VFGNN model. Additionally, to enhance optimization efficiency, we introduce a hyperparameter \(\epsilon\) as a trigger threshold. By appropriately setting this threshold, we ensure the stealthiness of the trigger while simplifying the optimization task from a triple optimization to a dual optimization.
We use the PGD method \cite{DBLP:conf/iclr/MadryMSTV18,huang2021unlearnable} to generate the trigger. The optimization formula for the trigger is as follows:
\begin{equation}\label{eq3}
{\delta}_{t+1}=\Pi_\epsilon\left(\delta_t-\alpha \cdot \operatorname{sgn} \left(  \nabla_{\delta} \mathcal{L}\left(F\left(a(\mathcal{G}_{p}, {\delta}_t); \Theta^*\right), \tau\right)\right)\right), 
\end{equation}
where \( t \) is the step index, \( \mathcal{G}_{p} \) represents the computation graph of the node \( v_p \) to inject the trigger, where \( v_p \in V_p \). \( \nabla_{\delta} \mathcal{L}\left(F\left(a(\mathcal{G}_{p}, {\delta}_t); \Theta^*\right), \tau\right) \) denotes the gradient of the loss function for the backdoor target class with respect to the trigger, \( \alpha \) is the step size, \( \Pi_\epsilon \) keeps \( \delta \) within an \(\epsilon\)-ball at each step, \( F(\Theta^*) \) refers to the pre-trained VFGNN model. \( \operatorname{sgn}(\cdot) \) denotes the sign function.

The multi-hop trigger \(\delta\) affects the target node and its neighbors, \(\delta = \{\delta^0, \delta^1, \cdots, \delta^M\}\). Here, \(\delta^m\) is the trigger added to the attributes of the \(m\)-hop neighbors of the target node \(v_p\). Therefore, the operation of adding triggers to \(\mathcal{G}_{p}\) can be expressed as
\begin{equation} \label{eq4}
a(\mathcal{G}_{p}, \delta) = (\mathbf{x}_p + \delta^0, \mathbf{X}_{1-hop} + \delta^1, \cdots, \mathbf{X}_{M-hop} + \delta^M),
\end{equation}
where \(\mathbf{x}_p\) is the attribute of node \(v_p\), \(\mathbf{X}_{m-hop}\) are the attributes of the \(m\)-hop neighbors, and \(\delta^0\), \(\delta^m\) are the triggers added to these attributes, respectively.

According to \eqref{eq3}, the generation of the trigger relies on the gradients of the target class nodes \(\tau\). The generation of the trigger and the training of the VFGNN model are accomplished together, which can be expressed through the following bi-level optimization:
\begin{equation} \label{eq5}
\begin{aligned}
\min_{\delta}\sum_{v_{i}\in\mathcal{V}_{P}}&\mathcal{L}(F(a(\mathcal{G}_{i},\delta);\Theta^{*}),\tau)
\\ \text{s.t.} \quad\Theta^{*}= \arg\min_{\Theta}&\sum_{v_{i}\in\mathcal{V}_{L}-\mathcal{V}_{P}}\mathcal{L}(F(G_{i};\Theta),y_{i})
\\+&\sum_{v_{i}\in\mathcal{V}_{P}}\mathcal{L}(F(a(G_{i},\delta);\Theta),\tau),
\end{aligned}
\end{equation}
where the trigger \(\delta\) is trained according to all known target class nodes \(\mathcal{V}_P\) available to the adversary, \(\mathcal{V}_P \in \mathcal{V}_L\). Essentially, this trigger serves as a universal trigger for the target class \(\tau\)  \cite{shafahi2020universal,DBLP:journals/corr/abs-2204-05255}.

\begin{algorithm}[!t]
 	\caption{Multi-hop Trigger Generation}
 	\label{alg:trigger_generation}
 	\begin{algorithmic}[1]
 		\REQUIRE{Training dataset $\mathcal{G}$; number of MTG epochs $T$; target class nodes $\mathcal{V}_P$ with target label $\tau$}
 		\ENSURE{ trigger $\delta$, model parameters $\Theta$} 
 		\STATE  Initialize: $\delta \leftarrow 0$.
        \WHILE{not reached $T$}
            \FOR{$v_i$ in $\mathcal{V}_L$ parallel}
                \STATE \textbf{Adversary} $A$: updates $\{\mathcal{G}_{i}^A = a(\mathcal{G}_{i}^A, {\delta})\}_{v_i \in \mathcal{V}_P}$.
                \STATE  \textbf{Passive party}:
                \FOR{each party $k=1,2,\dots, K$ parallel}
                \STATE $k$ computes embedded features $H_{i}^{k}$ using its bottom model $f_k$.
                \ENDFOR
                \STATE  \textbf{Active party}:
                \STATE  computes Eq. \eqref{eq1}, then updates $\theta_{Top}$ using $\frac{\partial \mathcal{L}}{ \partial \theta_{Top}}$.
                \STATE  sends $\frac{\partial \mathcal{L}}{\partial {H}_{i}}$ to all parties. 
                \STATE \textbf{Adversary} $A$:  \STATE computes $\nabla_{\delta} \mathcal{L}$ with  $\{\frac{\partial \mathcal{L}}{\partial {H}_{i}} \frac{\partial {H}_i}{\partial \delta}\}_{v_i \in \mathcal{V}_P}$\\
                \STATE updates $\delta$ with Eq. \eqref{eq3} 
                \STATE  \textbf{Passive party}:
                \FOR{each party $k=1,2,\dots, K$ parallel}
                \STATE $k$ computes $\nabla_{\theta_{k}} \mathcal{L}=\frac{\partial\mathcal{L}}{\partial H_{i} }\frac{\partial H_{i}^{k}}{\partial \theta_{k}}$. 
                \STATE $k$ updates model parameters $\theta_{k}$.
                \ENDFOR 
                
        \ENDFOR
        \ENDWHILE
 	\end{algorithmic}
 \end{algorithm}

Algorithm \ref{alg:trigger_generation} outlines the procedure of the proposed attack. In each iteration of VFGNN model training, the adversary A injects the trigger \(\delta\) into the local computation graph of known target class samples \(\mathcal{V}_P\) according to \eqref{eq4} (line 4). Subsequently, each participant \(k\) computes the node embeddings through the bottom model \(f_k\) (lines 5-8), where \(H_{i}^k\) represents the embedded features of the \(i\)-th data from the \(k\)-th participant. Upon receiving these embedded features, the active party computes the gradients of the loss function with respect to the top model and the embedded features of each participant according to \eqref{eq1}. The top model is then updated, and the gradients of the loss with respect to the embedded features \(\frac{\partial \mathcal{L}}{\partial {H}_i}\) are transmitted to each participant (lines 9-11), where \(H_i\) denotes the aggregation of the embedded features from all participants for the \(i\)-th data. After receiving the gradients, the adversary computes \(\nabla_{\delta} \mathcal{L}\) and updates the trigger using \eqref{eq3}. The current model parameters serve as the pre-trained model \(F(\Theta^*)\) (lines 12-14). Finally, each participant updates its bottom model based on the gradients sent by the active party (lines 15-19).

\subsection{Backdoor Retention}

When injecting backdoors, using only a small number of samples may lead to unstable effects. For instance, a backdoor may perform well in one epoch but degrade suddenly in the next. To ensure the stability of backdoors, we propose a method called Backdoor Retention (BR).

The BM method consists of two key steps. First, the attacker evaluates the impact of the backdoor. Since the attacker cannot directly access the output of the VFGNN model during training, they rely on an assumption: if the backdoor is successfully injected, the outputs \(H\) of nodes triggered by the injection should exhibit high similarity. Based on this assumption, the backdoor effectiveness \(E\) is approximated using the following formula:
\begin{equation}\label{eq:Effect}
E = \frac{1}{n^2} \sum_{i=1}^{n} \sum_{j=1}^{n} \frac{H_i \cdot H_j}{\|H_i\|_2 \|H_j\|_2},
\end{equation}
where \(n\) is the number of nodes involved in the trigger injection, and \(H_i\) represents the output of node \(v_i\) in the attacker's bottom model.

After estimating \(E\), the attacker performs the following actions in each epoch: if \(E\) exceeds a predefined threshold, the attacker updates the bottom model and trigger; otherwise, the model and trigger from the previous epoch are retained. This approach ensures the stability and effectiveness of the backdoor throughout the training process.

\section{Experiments} \label{experiment}

\subsection{Experiments Settings}

\subsubsection{Local GNN Models}
To demonstrate the effectiveness of BVG in VFGNN settings with different local GNN structures, we use three common GNN models as local participants:

\textbf{Graph Convolutional Network (GCN)} \cite{kipf2016semi}: GCN captures local features by propagating and aggregating information between nodes and their neighbors. Each local GNN model in VFGNN uses a two-layer GCN.
  
\textbf{Graph Attention Network (GAT)} \cite{velivckovic2017graph}: GAT uses an attention mechanism to assign adaptive weights to neighbor nodes, enhancing the model's ability to focus on important nodes.

\textbf{Simple Graph Convolution (SGC)} \cite{wu2019simplifying}: SGC simplifies GCN by removing nonlinear activations and reducing layers, making it more efficient while retaining essential graph structural information.

To ensure a fair comparison with previous studies \cite{chen2022graph}, we use a two-layer GNN model for each participant's local GNN to extract local node embeddings, with the dimension set to 16. The number of hidden units is fixed at 32. For GCN and GAT, the activation function is ReLU. VFGNN is trained using Adam, with a learning rate of 0.01. We provide a detailed description of the experimental details in the appendix \ref{appendix:expDetails}.



\subsubsection{Datasets}

This paper uses three widely adopted public datasets to evaluate the performance of BVG, including Cora \cite{mccallum2000automating}, Cora\_ml \cite{mccallum2000automating}, and Pubmed \cite{sen2008collective}. 
The basic dataset statistics are summarized in Table \ref{table:datasets}.

Each participant in the VFGNN framework has access to all the nodes in the datasets, but the node features are equally split among the participants. We randomly split the edges of the graphs into equal parts, one for each participant, without overlapped edges between any two participants. We assume the adversary knows only four target class nodes (i.e., $|\mathcal{V}_p|=4$), which are randomly selected from the training set.

\begin{table}[tbp]
\centering

\begin{tabular}{lcccc}
\toprule
\textbf{Datasets} & \textbf{Nodes} & \textbf{Edges} & \textbf{Features} & \textbf{Classes} \\
\midrule
Cora & 2708 & 5429 & 1433 & 7 \\
Cora\_ml & 2810 & 7981 & 2879 & 7 \\
Pubmed & 19717 & 44325 & 500 & 3 \\

\bottomrule
\end{tabular}
\caption{Basic information of the three datasets}
\label{table:datasets}
\end{table}

\subsubsection{Performance Metrics}
To evaluate the performance of the BVG method, we use three common metrics: Attack Success Rate (ASR), Main Task Accuracy (MTA) \cite{9802938}, and Mean Squared Error (MSE). Among them, ASR and MTA are the most commonly used evaluation metrics for backdoor attack tasks. ASR measures the proportion of samples in the backdoor test set that are predicted as the target class by the poisoned model. On the other hand, MTA assesses the accuracy of the clean test set on the poisoned model. MSE is used to measure the difference in the output of the bottom GNN model before and after trigger injection. A smaller MSE indicates that the injected trigger is less detectable by the active party, implying better stealth.

\begin{table*}[tbp]
    \centering
    \resizebox{0.90\textwidth}{!}{
    \begin{tabular}{ccccccccc}
        \toprule
        \midrule
        \textbf{Bottom} & \multirow{2}{*}{\textbf{Datasets}} & \multicolumn{7}{c}{\textbf{MTA}} \\
        \cmidrule(lr){3-9}
        \textbf{Model} &  & \textbf{GF$^*$} & \textbf{TECB} & \textbf{VILLAIN} & \textbf{ET} & \textbf{NFT} & \textbf{NFR} & \textbf{BVG} \\  
        \midrule
        \multirow{3}{*}{GCN} & Cora      & 48.60$\pm$2.87 & 62.72$\pm$4.67 & 65.38$\pm$2.99 & 65.28$\pm$2.37 & \bm{$66.12\pm2.13$} & 50.34$\pm$16.06 & 65.26$\pm$1.50 \\
                     & Cora\_ml  & 64.20$\pm$7.63 & \bm{$77.06\pm0.91$} & 73.60$\pm$3.49 & 75.50$\pm$2.78 & 75.46$\pm$2.25 & 56.90$\pm$21.80 & 76.50$\pm$2.63 \\
                     & Pubmed    & 44.00$\pm$10.20 & 64.90$\pm$4.15 & 69.26$\pm$2.86 & 59.80$\pm$15.40 & 68.44$\pm$2.58 & 32.36$\pm$13.57 &  \bm{$71.34\pm3.54$} \\
        \midrule
        \multirow{3}{*}{GAT} & Cora      & 46.60$\pm$3.93 & 40.06$\pm$4.10 & 63.44$\pm$2.01 & 61.70$\pm$6.37 &  \bm{$64.30\pm3.04$} & 41.02$\pm$16.66 & 63.72$\pm$2.20 \\
                     & Cora\_ml  & 61.00$\pm$3.29 & 57.78$\pm$4.86 & 73.14$\pm$1.75 & 76.12$\pm$2.45 &  \bm{$76.50\pm2.12$} & 59.94$\pm$23.23 & 75.58$\pm$2.77 \\
                     & Pubmed    & 70.80$\pm$3.54 & 44.98$\pm$17.92 & 72.40$\pm$0.89 & 63.66$\pm$11.79 & 72.88$\pm$1.49 & 35.64$\pm$14.71 &  \bm{$73.28\pm1.73$} \\
        \midrule
        \multirow{3}{*}{SGC} & Cora      & 49.00$\pm$3.03 & 36.68$\pm$29.97 & 62.56$\pm$2.12 & 55.04$\pm$16.96 & 62.08$\pm$2.63 & 50.12$\pm$15.24 &  \bm{$65.30\pm2.08$} \\
                     & Cora\_ml  & 67.40$\pm$4.03 & 74.76$\pm$2.62 & 74.08$\pm$2.47 &  \bm{$75.34\pm1.62$} & 75.08$\pm$3.59 & 56.80$\pm$21.77 & 74.94$\pm$2.39 \\
                     & Pubmed    & 41.82$\pm$12.33 & 54.04$\pm$27.02 & 68.90$\pm$4.07 & 61.46$\pm$14.75 & 69.80$\pm$1.94 & 39.02$\pm$11.83 & \bm{$70.64\pm2.95$} \\
        \midrule
         \textbf{Bottom} & \multirow{2}{*}{\textbf{Datasets}} & \multicolumn{7}{c}{\textbf{ASR}} \\
        \cmidrule(lr){3-9}
        \textbf{Model} &  & \textbf{GF} & \textbf{TECB} & \textbf{VILLAIN} & \textbf{ET} & \textbf{NFT} & \textbf{NFR} & \textbf{BVG} \\  
        \midrule
        \multirow{3}{*}{GCN} & Cora & 51.40$\pm$5.61 & 16.40$\pm$10.91 & 99.00$\pm$0.64 & 48.80$\pm$13.62 & 91.54$\pm$4.13 & 36.48$\pm$22.01 & \bm{$99.86\pm0.28$} \\
        & Cora\_ml & 26.80$\pm$5.15 & 34.60$\pm$7.74 & 97.34$\pm$3.94 & 49.92$\pm$7.82 & 74.78$\pm$22.86 & 38.12$\pm$31.24 & \bm{$99.98\pm0.04$} \\
        & Pubmed & 24.00$\pm$8.00 & 81.40$\pm$13.83 & 94.96$\pm$4.44 & 78.96$\pm$8.91 & 88.02$\pm$6.41 & 81.86$\pm$20.90 & \bm{$100.00\pm0.00$} \\
        \midrule
        \multirow{3}{*}{GAT} & Cora & 15.80$\pm$6.31 & 56.80$\pm$35.54 & 94.52$\pm$5.64 & 52.00$\pm$14.36 & 66.14$\pm$28.01 & 43.34$\pm$30.03 & \bm{$99.12\pm1.10$} \\
        & Cora\_ml & 23.40$\pm$2.73 & 45.14$\pm$31.03 & 90.48$\pm$6.33 & 44.80$\pm$17.86 & 95.28$\pm$4.82 & 34.62$\pm$32.96 & \bm{$99.92\pm0.16$} \\
        & Pubmed & 30.20$\pm$2.04 & 89.60$\pm$20.80 & 86.98$\pm$10.71 & 78.42$\pm$9.90 & 98.64$\pm$2.23 & 80.62$\pm$20.43 & \bm{$100.00\pm0.00$} \\
        \midrule
        \multirow{3}{*}{SGC} & Cora & 51.00$\pm$4.20 & 12.24$\pm$13.28 & 99.34$\pm$0.65 & 54.40$\pm$22.44 & 87.60$\pm$13.96 & 32.46$\pm$19.15 & \bm{$99.76\pm0.39$} \\
        & Cora\_ml & 27.80$\pm$3.97 & 15.18$\pm$9.14 & 98.72$\pm$1.42 & 60.22$\pm$12.20 & 88.66$\pm$7.56 & 37.78$\pm$31.25 & \bm{$99.98\pm0.04$} \\
        & Pubmed & 32.73$\pm$10.91 & 59.78$\pm$48.81 & 99.10$\pm$0.94 & 82.56$\pm$5.47 & 91.30$\pm$6.20 & 71.84$\pm$15.77 & \bm{$99.90\pm0.20$} \\
        
        \midrule
        \bottomrule
    \end{tabular}
    }
    \caption{Performance comparison with other methods on three datasets and three bottom models (Standard Deviation Included, Best Results Highlighted in Bold. GF$^*$ is an adversarial attack method, where a lower MTA is desirable.)}
    \label{tab:performance_comparison}
\end{table*}

\subsubsection{Comparison Benchmarks}
Due to the lack of research on backdoor attacks in VFGNN, we selected three attack methods from other scenarios for comparison, adapting them to fit VFGNN's tasks and structure. Additionally, we incorporated common GNN backdoor trigger schemes for further comparison. The methods are summarized below:

\textbf{GF} \cite{chen2022graph}: Originally designed for adversarial tasks in VFGNN, this method was adapted to minimize the loss of the backdoor class. It involves querying participant embeddings, training a proxy model, and generating an adversarial graph on the proxy model.

\textbf{TECB} \cite{chen2023practical}: A VFL backdoor attack method for image data, adapted here for graph tasks by aligning target node attributes with trigger embeddings. It operates in two stages: training triggers with limited labels and aligning target gradients using the trigger.

\textbf{VILLAIN} \cite{bai2023villain}: Designed for VFL backdoor attacks, this method injects triggers into the bottom model's output  \(H\), supporting clean-label attacks. It combines label inference using limited labeled samples and data poisoning with trigger masking and randomness for enhanced robustness.

\textbf{Edge Triggering (ET)}: This method connects attack nodes to target class nodes via edges, creating a trigger relationship that exploits the graph's topology to influence target node classification.

\textbf{Node Feature Triggering (NFT)}: Backdoor attacks are achieved by injecting trigger patterns into nodes with known labels, altering their features to misclassify attack nodes as the target class under specific triggers.

\textbf{Node Feature Replacement (NFR)}: Attackers replace the features of attack nodes with those of target class nodes, making their features similar to the target class and inducing misclassification.

\subsection{Experiment Results Analysis}

\subsubsection{Backdoor Task Evaluation} 

We conducted experiments in a VFGNN framework with two participants, as shown in Table \ref{tab:performance_comparison}. BVG and NFT perform best on the main task, with slight variations across datasets and models. For the backdoor task, BVG consistently achieves the highest attack success rate across all tested models and datasets, demonstrating its effectiveness with minimal impact on main task performance.


\begin{figure}[!tbp]
\centerline{\includegraphics[width=\linewidth]{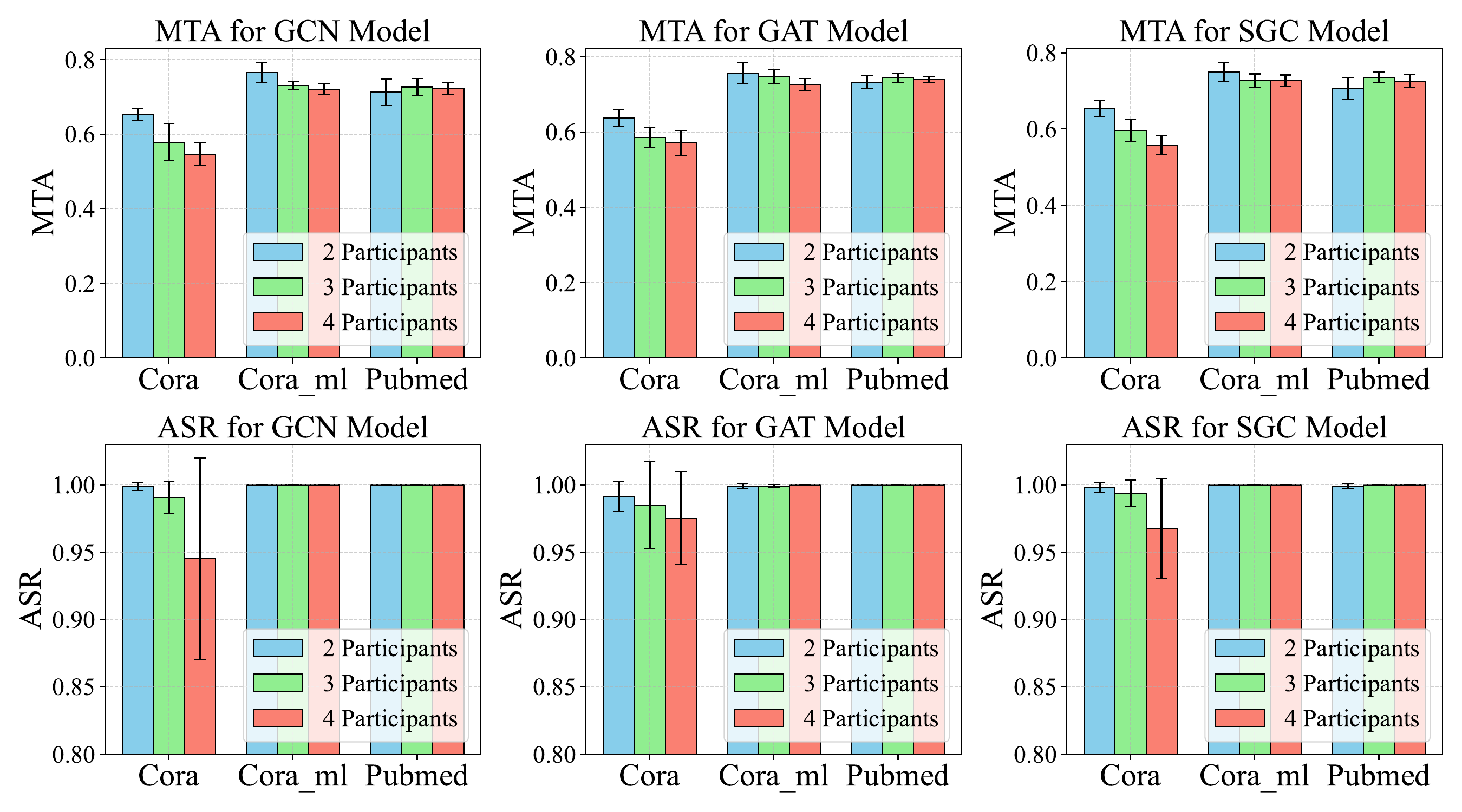}}
\caption{Performance of BVG under Multi-Party Settings (Standard Deviation Included.)}
\label{fig:participants}
\end{figure}

We also evaluated the proposed attack method's performance in multi-party VFGNN scenarios where multiple participants are involved. Unlike the two-party setting, multi-party scenarios become more complex due to the increased number of participants and their interactions. Our evaluation focuses on how the presence of multiple participants affects the ASR and the MTA of BVG.

We conducted experiments with two, three, and four participants to evaluate the scalability and effectiveness of the proposed backdoor attack method in multi-party VFGNN scenarios. Each participant was assigned a subset of the graph's edges and node features, with one active party holding the labels for the labeled node set $V_L$. The passive participants, including the adversary, only had partial information about the graph. The experimental results are shown in Figure \ref{fig:participants}.

The results indicate that the main task accuracy slightly decreases as the number of participants increases. This trend could be attributed to the increased complexity and noise introduced by multiple participants. The backdoor attack success rate remains high across all multi-party settings but slightly decreases as the number of participants increases. The obtained results verify that our attack method is effective even in more complex multi-party environments. The slight decrease in success rate may be due to the dilution of the adversary's influence with more participants.

\subsubsection{Backdoor Stealthiness Evaluation} 

\begin{figure}[!tbp]
\centerline{\includegraphics[width=\linewidth]{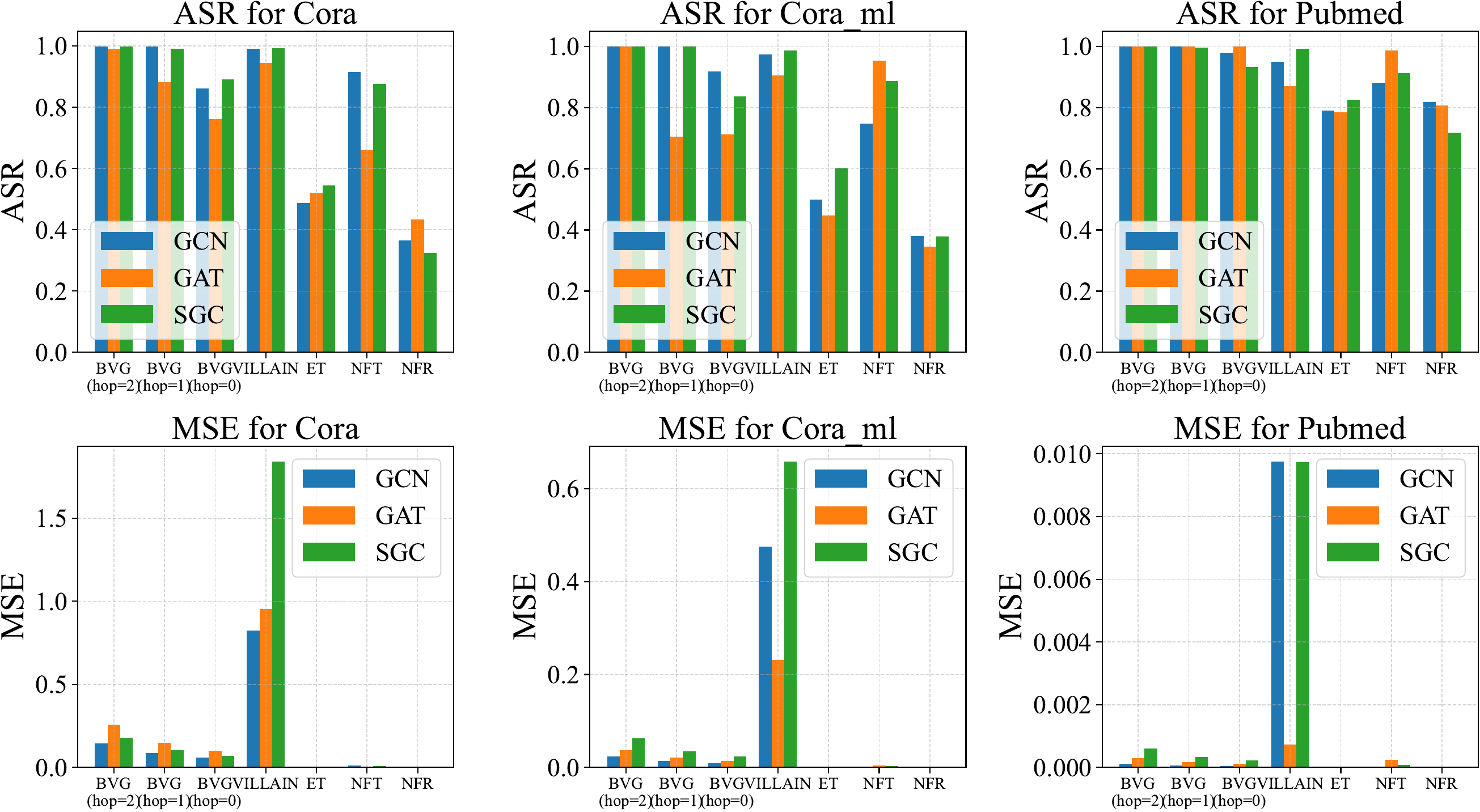}}
\caption{Comparison of Backdoor Task Performance and Intermediate Layer Feature Differences Across Methods}
\label{fig:mse_comparison}
\end{figure}
In evaluating the stealthiness of backdoors in VFGNN, we measure the difference in the intermediate layer features of the same node before and after the backdoor injection. If the intermediate layer features show little change after the backdoor injection, we consider the backdoor difficult for the attacker to detect. For comparison, we selected several methods from Table \ref{tab:performance_comparison} that perform relatively well on the backdoor task, with a particular focus on the performance of the BVG method under multi-hop triggers. Figure 4 shows the backdoor task performance and the average difference in intermediate layer features for five methods across three datasets.

From Figure \ref{fig:mse_comparison}, it is evident that the increase in MSE correlates positively with the increase in ASR. However, despite BVG achieving a significantly higher ASR than VILLAIN, its MSE is much lower than VILLAIN, indicating that BVG exhibits better stealthiness. For the other methods, BVG has slightly worse stealthiness but significantly better ASR performance. More importantly, the ASR performance of ET, NFT, and NFR has almost plateaued, as discussed in Appendix \ref{ComparisonMethods}.



\subsubsection{Attack Efficacy under Defense} 

To evaluate the robustness of the BVG method, we explored potential backdoor defense strategies within the VFGNN framework. We considered two main types of defense: the first type is GNN backdoor defense methods, such as Prune and Prune+LD \cite{dai2023unnoticeable}. These methods defend against backdoor attacks by pruning the edges between nodes with low similarity. Pruning these edges can disrupt the attacker's trigger structure and connections. However, VFGNN's architecture prevents defenders from pruning the adversary's data, making these methods unsuitable. 


The second type of defense methods is specific to the VFL framework and can be divided into two categories: label inference defense and backdoor attack defense. Label inference defense operates on the assumption that backdoor attacks often require access to a large amount of label information. To prevent label leakage, defense methods perturb or compress gradients to minimize information leakage. Two representative methods are DP-SGD and gradient compression (GC). DP-SGD provides differential privacy protection by adding noise to gradients \cite{277244}, where the larger the noise amplitude, the smaller the privacy leakage, and vice versa. The GC method \cite{277244,kairouz2021advances} sends only a subset of gradients with the largest absolute values to participants, thereby reducing privacy leakage. For backdoor attack defense, there are mainly two approaches: one is model reconstruction, which aims to purify the top model of the active party to remove the backdoor, with methods such as model pruning \cite{liu2018fine} and adversarial neuron pruning (ANP) \cite{wu2021adversarial}. The other approach is to introduce interference into intermediate-layer features \cite{li2021backdoor}, such as embedding Gaussian noise into the intermediate features provided to the passive party to disrupt potential backdoor triggers.

\begin{figure}[!tbp]
\centerline{\includegraphics[width=\linewidth]{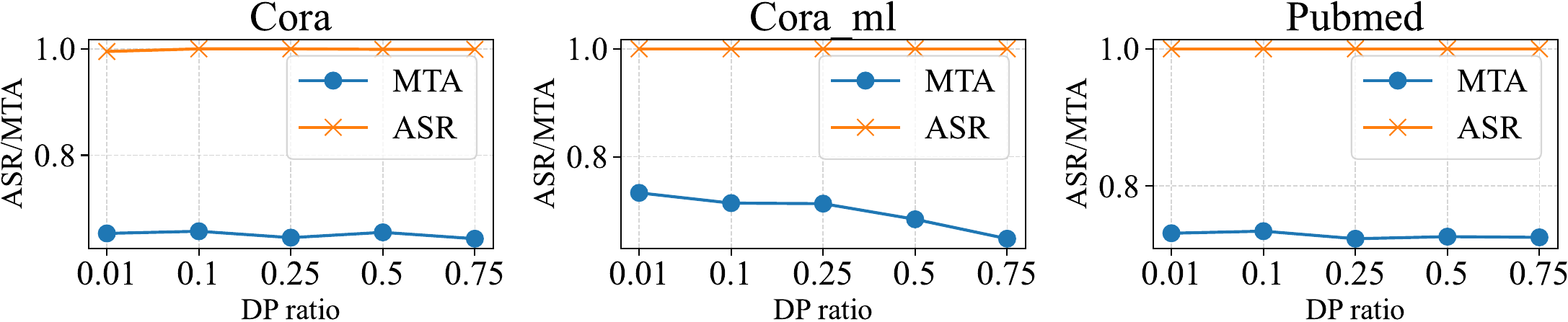}}
\caption{The performance of BVG under DP-SGD defense}
\label{fig:DP_gcn}
\end{figure}

\begin{figure}[!tbp]
\centerline{\includegraphics[width=\linewidth]{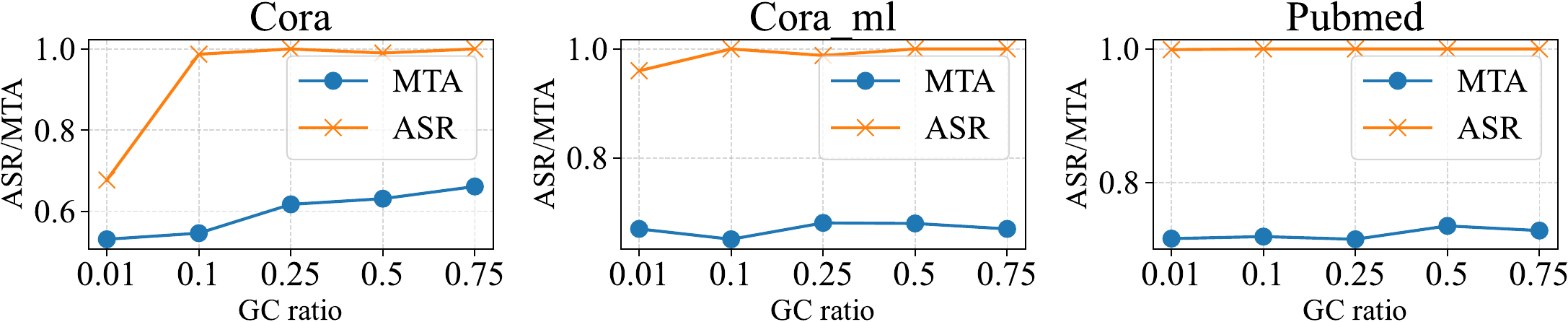}}
\caption{The performance of BVG under GC defense}
\label{fig:GC_gcn}
\end{figure}

\begin{figure}[!tbp]
\centerline{\includegraphics[width=\linewidth]{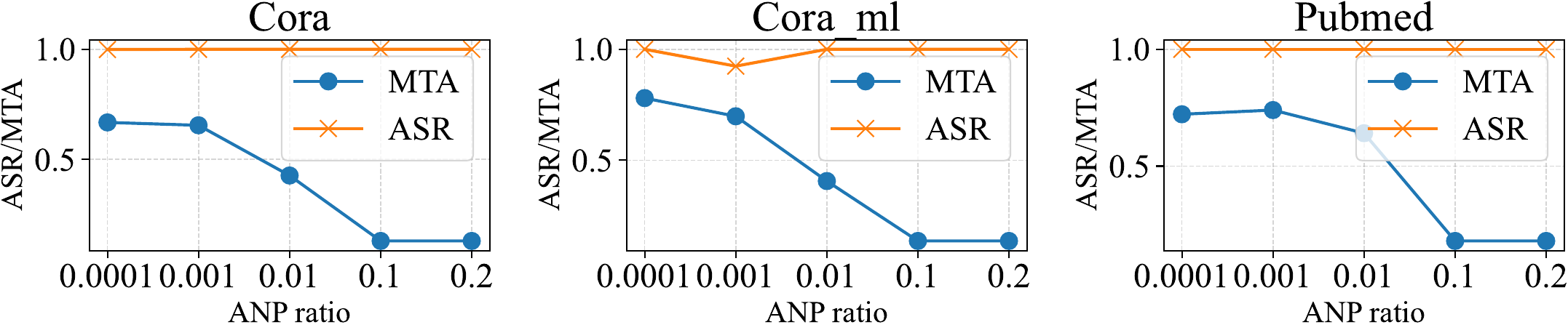}}
\caption{The performance of BVG under ANP defense}
\label{fig:ANP_gcn}
\end{figure}
\begin{figure}[!tbp]
\centerline{\includegraphics[width=\linewidth]{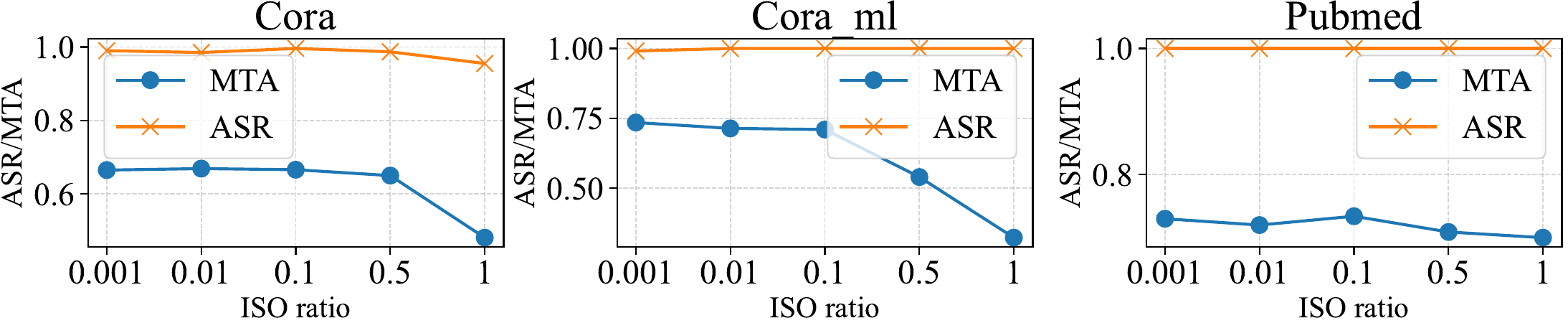}}
\caption{The performance of BVG under ISO defense}
\label{fig:ISO_gcn}
\end{figure}

In the VFGNN framework, we implemented several VFL backdoor defense methods, each with five parameter settings, as shown in the figure.
We present the experimental results when the base model is GCN, with additional results provided in Appendix \ref{appendix:defen}. For label inference defense, as shown in Figures \ref{fig:DP_gcn} and \ref{fig:GC_gcn}, our method is minimally affected, as the BVG method itself does not rely on label inference. 
For backdoor attack defense, the results of the ANP method are shown in Figure \ref{fig:ANP_gcn}. 
The ANP method primarily affects model accuracy without effectively suppressing the backdoor. For the defense method involving interference with intermediate-layer features, the experimental results are shown in Figure \ref{fig:ISO_gcn}. As model accuracy decreases, the success rate of backdoor attacks also drops, making it difficult for defenders to reduce ASR while maintaining a high MTA.

\section{Conclusion}\label{conclusion}

In this paper, we proposed the BVG method for ckdoor attacks to Vertical Federated Graph Neural Network (VFGNN). Utilizing a multi-hop trigger generation approach, the BVG method can perform efficient backdoor attacks with very limited knowledge of target class nodes. 
Extensive experiments on three datasets with three GNN models demonstrate that BVG achieves a high attack success rate while having a minimal impact on the main task accuracy and is unlikely to be detected by the active party.
The evaluation of BVG efficacy under various defense methods highlights the robustness and efficiency of the proposed attack method, underscoring the necessity for advanced defense mechanisms in practical federated learning applications to counter such sophisticated backdoor attacks.

\bibliographystyle{named}
\bibliography{ijcai25}

\begin{thebibliography}{}

\bibitem[\protect\citeauthoryear{Bai \bgroup \em et al.\egroup }{2023}]{bai2023villain}
Yijie Bai, Yanjiao Chen, Hanlei Zhang, Wenyuan Xu, Haiqin Weng, and Dou Goodman.
\newblock $\{$VILLAIN$\}$: Backdoor attacks against vertical split learning.
\newblock In {\em 32nd USENIX Security Symposium (USENIX Security 23)}, pages 2743--2760, 2023.

\bibitem[\protect\citeauthoryear{Chen \bgroup \em et al.\egroup }{2020}]{chen2020vertically}
Chaochao Chen, Jun Zhou, Longfei Zheng, Huiwen Wu, Lingjuan Lyu, Jia Wu, Bingzhe Wu, Ziqi Liu, Li~Wang, and Xiaolin Zheng.
\newblock Vertically federated graph neural network for privacy-preserving node classification.
\newblock {\em arXiv preprint arXiv:2005.11903}, 2020.

\bibitem[\protect\citeauthoryear{Chen \bgroup \em et al.\egroup }{2022}]{chen2022graph}
Jinyin Chen, Guohan Huang, Haibin Zheng, Shanqing Yu, Wenrong Jiang, and Chen Cui.
\newblock Graph-fraudster: Adversarial attacks on graph neural network-based vertical federated learning.
\newblock {\em IEEE Transactions on Computational Social Systems}, 10(2):492--506, 2022.

\bibitem[\protect\citeauthoryear{Chen \bgroup \em et al.\egroup }{2023a}]{chen2023practical}
Peng Chen, Jirui Yang, Junxiong Lin, Zhihui Lu, Qiang Duan, and Hongfeng Chai.
\newblock A practical clean-label backdoor attack with limited information in vertical federated learning.
\newblock In {\em 2023 IEEE International Conference on Data Mining (ICDM)}, pages 41--50. IEEE, 2023.

\bibitem[\protect\citeauthoryear{Chen \bgroup \em et al.\egroup }{2023b}]{chen2023feature}
Yang Chen, Zhonglin Ye, Haixing Zhao, and Ying Wang.
\newblock Feature-based graph backdoor attack in the node classification task.
\newblock {\em International Journal of Intelligent Systems}, 2023(1):5418398, 2023.

\bibitem[\protect\citeauthoryear{Chen \bgroup \em et al.\egroup }{2024}]{chen2024universal}
Peng Chen, Xin Du, Zhihui Lu, and Hongfeng Chai.
\newblock Universal adversarial backdoor attacks to fool vertical federated learning.
\newblock {\em Computers \& Security}, 137:103601, 2024.

\bibitem[\protect\citeauthoryear{Cheng \bgroup \em et al.\egroup }{2023}]{cheng2023anti}
Dawei Cheng, Yujia Ye, Sheng Xiang, Zhenwei Ma, Ying Zhang, and Changjun Jiang.
\newblock Anti-money laundering by group-aware deep graph learning.
\newblock {\em IEEE Transactions on Knowledge and Data Engineering}, 35(12):12444--12457, 2023.

\bibitem[\protect\citeauthoryear{Dai \bgroup \em et al.\egroup }{2023}]{dai2023unnoticeable}
Enyan Dai, Minhua Lin, Xiang Zhang, and Suhang Wang.
\newblock Unnoticeable backdoor attacks on graph neural networks.
\newblock In {\em Proceedings of the ACM Web Conference 2023}, pages 2263--2273, 2023.

\bibitem[\protect\citeauthoryear{Fu \bgroup \em et al.\egroup }{2022}]{277244}
Chong Fu, Xuhong Zhang, Shouling Ji, Jinyin Chen, Jingzheng Wu, Shanqing Guo, Jun Zhou, Alex~X Liu, and Ting Wang.
\newblock Label inference attacks against vertical federated learning.
\newblock In {\em 31st USENIX Security Symposium (USENIX Security 22)}, pages 1397--1414, 2022.

\bibitem[\protect\citeauthoryear{Gu and Bai}{2023}]{DBLP:journals/compsec/GuB23}
Yuhao Gu and Yuebin Bai.
\newblock Lr-ba: Backdoor attack against vertical federated learning using local latent representations.
\newblock {\em Computers \& Security}, 129:103193, 2023.

\bibitem[\protect\citeauthoryear{He \bgroup \em et al.\egroup }{2021}]{he2021fedgraphnn}
Chaoyang He, Keshav Balasubramanian, Emir Ceyani, Carl Yang, Han Xie, Lichao Sun, Lifang He, Liangwei Yang, Philip~S Yu, Yu~Rong, et~al.
\newblock Fedgraphnn: A federated learning system and benchmark for graph neural networks.
\newblock {\em arXiv preprint arXiv:2104.07145}, 2021.

\bibitem[\protect\citeauthoryear{He \bgroup \em et al.\egroup }{2023}]{he2023backdoor}
Ying He, Zhili Shen, Jingyu Hua, Qixuan Dong, Jiacheng Niu, Wei Tong, Xu~Huang, Chen Li, and Sheng Zhong.
\newblock Backdoor attack against split neural network-based vertical federated learning.
\newblock {\em IEEE Transactions on Information Forensics and Security}, 2023.

\bibitem[\protect\citeauthoryear{Huang \bgroup \em et al.\egroup }{2021}]{huang2021unlearnable}
Hanxun Huang, Xingjun Ma, Sarah~Monazam Erfani, James Bailey, and Yisen Wang.
\newblock Unlearnable examples: Making personal data unexploitable.
\newblock In {\em International Conference on Learning Representations}, 2021.

\bibitem[\protect\citeauthoryear{Kairouz \bgroup \em et al.\egroup }{2021}]{kairouz2021advances}
Peter Kairouz, H~Brendan McMahan, Brendan Avent, Aur{\'e}lien Bellet, Mehdi Bennis, Arjun~Nitin Bhagoji, Kallista Bonawitz, Zachary Charles, Graham Cormode, Rachel Cummings, et~al.
\newblock Advances and open problems in federated learning.
\newblock {\em Foundations and Trends{\textregistered} in Machine Learning}, 14(1--2):1--210, 2021.

\bibitem[\protect\citeauthoryear{Kipf and Welling}{2016}]{kipf2016semi}
Thomas~N Kipf and Max Welling.
\newblock Semi-supervised classification with graph convolutional networks.
\newblock {\em arXiv preprint arXiv:1609.02907}, 2016.

\bibitem[\protect\citeauthoryear{Li \bgroup \em et al.\egroup }{2021}]{li2021backdoor}
Yiming Li, Tongqing Zhai, Yong Jiang, Zhifeng Li, and Shu-Tao Xia.
\newblock Backdoor attack in the physical world.
\newblock {\em arXiv preprint arXiv:2104.02361}, 2021.

\bibitem[\protect\citeauthoryear{Li \bgroup \em et al.\egroup }{2022a}]{li2022backdoor}
Yiming Li, Yong Jiang, Zhifeng Li, and Shu-Tao Xia.
\newblock Backdoor learning: A survey.
\newblock {\em IEEE Transactions on Neural Networks and Learning Systems}, 2022.

\bibitem[\protect\citeauthoryear{Li \bgroup \em et al.\egroup }{2022b}]{9802938}
Yiming Li, Yong Jiang, Zhifeng Li, and Shu-Tao Xia.
\newblock Backdoor learning: A survey.
\newblock {\em IEEE Transactions on Neural Networks and Learning Systems}, 2022.

\bibitem[\protect\citeauthoryear{Liu \bgroup \em et al.\egroup }{2018}]{liu2018fine}
Kang Liu, Brendan Dolan-Gavitt, and Siddharth Garg.
\newblock Fine-pruning: Defending against backdooring attacks on deep neural networks.
\newblock In {\em International symposium on research in attacks, intrusions, and defenses}, pages 273--294. Springer, 2018.

\bibitem[\protect\citeauthoryear{Liu \bgroup \em et al.\egroup }{2022}]{DBLP:conf/nips/LiuXK022}
Jing Liu, Chulin Xie, Sanmi Koyejo, and Bo~Li.
\newblock Copur: Certifiably robust collaborative inference via feature purification.
\newblock {\em Advances in Neural Information Processing Systems}, 35:26645--26657, 2022.

\bibitem[\protect\citeauthoryear{Liu \bgroup \em et al.\egroup }{2023}]{liu2023bkd}
Fan Liu, Siqi Lai, Yansong Ning, and Hao Liu.
\newblock Bkd-fedgnn: A benchmark for classification backdoor attacks on federated graph neural network.
\newblock {\em arXiv preprint arXiv:2306.10351}, 2023.

\bibitem[\protect\citeauthoryear{Liu \bgroup \em et al.\egroup }{2024}]{FedGNN-survey-TNNLS24}
Rui Liu, Pengwei Xing, Zichao Deng, Anran Li, Cuntai Guan, and Han Yu.
\newblock Federated graph neural networks: Overview, techniques, and challenges.
\newblock {\em IEEE Transactions on Neural Networks and Learning Systems}, pages 1--17, 2024.

\bibitem[\protect\citeauthoryear{Madry \bgroup \em et al.\egroup }{2018}]{DBLP:conf/iclr/MadryMSTV18}
Aleksander Madry, Aleksandar Makelov, Ludwig Schmidt, Dimitris Tsipras, and Adrian Vladu.
\newblock Towards deep learning models resistant to adversarial attacks.
\newblock In {\em International Conference on Learning Representations}, 2018.

\bibitem[\protect\citeauthoryear{Mai and Pang}{2023}]{mai2023vertical}
Peihua Mai and Yan Pang.
\newblock Vertical federated graph neural network for recommender system.
\newblock In {\em International Conference on Machine Learning}, pages 23516--23535. PMLR, 2023.

\bibitem[\protect\citeauthoryear{McCallum \bgroup \em et al.\egroup }{2000}]{mccallum2000automating}
Andrew~Kachites McCallum, Kamal Nigam, Jason Rennie, and Kristie Seymore.
\newblock Automating the construction of internet portals with machine learning.
\newblock {\em Information Retrieval}, 3:127--163, 2000.

\bibitem[\protect\citeauthoryear{Naseri \bgroup \em et al.\egroup }{2024}]{naseri2023badvfl}
Mohammad Naseri, Yufei Han, and Emiliano De~Cristofaro.
\newblock Badvfl: Backdoor attacks in vertical federated learning.
\newblock In {\em 2024 IEEE Symposium on Security and Privacy (SP)}, pages 2013--2028. IEEE, 2024.

\bibitem[\protect\citeauthoryear{Sen \bgroup \em et al.\egroup }{2008}]{sen2008collective}
Prithviraj Sen, Galileo Namata, Mustafa Bilgic, Lise Getoor, Brian Galligher, and Tina Eliassi-Rad.
\newblock Collective classification in network data.
\newblock {\em AI magazine}, 29(3):93--93, 2008.

\bibitem[\protect\citeauthoryear{Shafahi \bgroup \em et al.\egroup }{2020}]{shafahi2020universal}
Ali Shafahi, Mahyar Najibi, Zheng Xu, John Dickerson, Larry~S Davis, and Tom Goldstein.
\newblock Universal adversarial training.
\newblock In {\em Proceedings of the AAAI Conference on Artificial Intelligence}, volume~34, pages 5636--5643, 2020.

\bibitem[\protect\citeauthoryear{Veli{\v{c}}kovi{\'c} \bgroup \em et al.\egroup }{2017}]{velivckovic2017graph}
Petar Veli{\v{c}}kovi{\'c}, Guillem Cucurull, Arantxa Casanova, Adriana Romero, Pietro Lio, and Yoshua Bengio.
\newblock Graph attention networks.
\newblock {\em arXiv preprint arXiv:1710.10903}, 2017.

\bibitem[\protect\citeauthoryear{Wang \bgroup \em et al.\egroup }{2024}]{wang2024explanatory}
Huiwei Wang, Tianhua Liu, Ziyu Sheng, and Huaqing Li.
\newblock Explanatory subgraph attacks against graph neural networks.
\newblock {\em Neural Networks}, 172:106097, 2024.

\bibitem[\protect\citeauthoryear{Wu and Wang}{2021}]{wu2021adversarial}
Dongxian Wu and Yisen Wang.
\newblock Adversarial neuron pruning purifies backdoored deep models.
\newblock {\em Advances in Neural Information Processing Systems}, 34:16913--16925, 2021.

\bibitem[\protect\citeauthoryear{Wu \bgroup \em et al.\egroup }{2019}]{wu2019simplifying}
Felix Wu, Amauri Souza, Tianyi Zhang, Christopher Fifty, Tao Yu, and Kilian Weinberger.
\newblock Simplifying graph convolutional networks.
\newblock In {\em International conference on machine learning}, pages 6861--6871. PMLR, 2019.

\bibitem[\protect\citeauthoryear{Wu \bgroup \em et al.\egroup }{2020}]{wu2020comprehensive}
Zonghan Wu, Shirui Pan, Fengwen Chen, Guodong Long, Chengqi Zhang, and S~Yu Philip.
\newblock A comprehensive survey on graph neural networks.
\newblock {\em IEEE transactions on neural networks and learning systems}, 32(1):4--24, 2020.

\bibitem[\protect\citeauthoryear{Xi \bgroup \em et al.\egroup }{2021}]{xi2021graph}
Zhaohan Xi, Ren Pang, Shouling Ji, and Ting Wang.
\newblock Graph backdoor.
\newblock In {\em 30th USENIX security symposium (USENIX Security 21)}, pages 1523--1540, 2021.

\bibitem[\protect\citeauthoryear{Xing \bgroup \em et al.\egroup }{2023}]{xing2023clean}
Xiaogang Xing, Ming Xu, Yujing Bai, and Dongdong Yang.
\newblock A clean-label graph backdoor attack method in node classification task.
\newblock {\em arXiv preprint arXiv:2401.00163}, 2023.

\bibitem[\protect\citeauthoryear{Xu \bgroup \em et al.\egroup }{2022}]{xu2022more}
Jing Xu, Rui Wang, Stefanos Koffas, Kaitai Liang, and Stjepan Picek.
\newblock More is better (mostly): On the backdoor attacks in federated graph neural networks.
\newblock In {\em Proceedings of the 38th Annual Computer Security Applications Conference}, pages 684--698, 2022.

\bibitem[\protect\citeauthoryear{Yang \bgroup \em et al.\egroup }{2022}]{yang2022transferable}
Shuiqiao Yang, Bao~Gia Doan, Paul Montague, Olivier De~Vel, Tamas Abraham, Seyit Camtepe, Damith~C Ranasinghe, and Salil~S Kanhere.
\newblock Transferable graph backdoor attack.
\newblock In {\em Proceedings of the 25th international symposium on research in attacks, intrusions and defenses}, pages 321--332, 2022.

\bibitem[\protect\citeauthoryear{Zeng \bgroup \em et al.\egroup }{2022}]{DBLP:journals/corr/abs-2204-05255}
Yi~Zeng, Minzhou Pan, Hoang~Anh Just, Lingjuan Lyu, Meikang Qiu, and Ruoxi Jia.
\newblock Narcissus: A practical clean-label backdoor attack with limited information.
\newblock {\em arXiv preprint arXiv:2204.05255}, 2022.

\bibitem[\protect\citeauthoryear{Zhang \bgroup \em et al.\egroup }{2021}]{GNN-backdoor-ccfc21}
Zaixi Zhang, Jinyuan Jia, Binghui Wang, and Neil~Zhenqiang Gong.
\newblock Backdoor attacks to graph neural networks.
\newblock In {\em Proceedings of the 26th ACM Symposium on Access Control Models and Technologies}, pages 15--26, 2021.

\bibitem[\protect\citeauthoryear{Zhang \bgroup \em et al.\egroup }{2024}]{zhang2024a3fl}
Hangfan Zhang, Jinyuan Jia, Jinghui Chen, Lu~Lin, and Dinghao Wu.
\newblock A3fl: Adversarially adaptive backdoor attacks to federated learning.
\newblock {\em Advances in Neural Information Processing Systems}, 36, 2024.

\bibitem[\protect\citeauthoryear{Zhao \bgroup \em et al.\egroup }{2020}]{zhao2020clean}
Shihao Zhao, Xingjun Ma, Xiang Zheng, James Bailey, Jingjing Chen, and Yu-Gang Jiang.
\newblock Clean-label backdoor attacks on video recognition models.
\newblock In {\em Proceedings of the IEEE/CVF Conference on Computer Vision and Pattern Recognition}, pages 14443--14452, 2020.

\bibitem[\protect\citeauthoryear{Zheng \bgroup \em et al.\egroup }{2023}]{zheng2023link}
Haibin Zheng, Haiyang Xiong, Haonan Ma, Guohan Huang, and Jinyin Chen.
\newblock Link-backdoor: Backdoor attack on link prediction via node injection.
\newblock {\em IEEE Transactions on Computational Social Systems}, 2023.

\end{thebibliography}

\newpage

\appendix

\section{Experimental Details}\label{appendix:expDetails}
Our experiments are conducted in a VFGNN framework with only two participants by default. One is the active party, responsible for providing labels, and the other is the passive party, which provides data to assist the active party in training. Compared to Horizontal Federated Learning (HFL), the number of participants in Vertical Federated Learning (VFL) is typically smaller. This is because VFL requires the participants to have a large number of overlapping sample IDs while their features are complementary. However, in practice, it is almost impossible for dozens of companies with nearly identical user groups to simultaneously agree to collaborate. Following the experimental setup in existing VFL research, we assume there are only two participants by default \cite{277244,bai2023villain}.

For nine groups of experiments involving three datasets and three base models, we set the batch size for each training batch to 64, and the trigger update magnitude to 0.2. For the six experiments conducted on Cora\_ml and Pubmed, we set the maximum trigger value \(\epsilon\) to 0.08. For the experiments on Cora, we set the maximum \(\epsilon\) to 0.1 for GCN and 0.08 for GAT and SGC. 

In the Backdoor Retention process, we first train the model normally for the first five epochs, and then implement Backdoor Retention starting from the sixth epoch. To evaluate the backdoor effect, we randomly select 10 points for testing, noting that these 10 points do not need to be labeled points. In the experiments using GCN as the bottom model, the thresholds for the Cora, Cora\_ml, and Pubmed datasets are set to 0.985, 0.98, and 0.98, respectively. In the experiments with GAT as the bottom model, the thresholds for the three datasets are set to 0.95, 0.98, and 0.98, respectively. For the SGC-based experiments, the thresholds for Cora, Cora\_ml, and Pubmed are set to 0.9, 0.96, and 0.98, respectively.

This translation keeps the technical details intact while ensuring clarity in academic writing.

The experiments were run on a server equipped with an Intel(R) Xeon(R) Silver 4310 CPU @ 2.10GHz and an NVIDIA GTX 3090 GPU. Each experiment in Table \ref{tab:performance_comparison} was conducted five times to evaluate the overall performance, and we recorded the average and standard deviation.


\subsection{Design of Comparison Methods}
To the best of our knowledge, there are currently no backdoor attack methods specifically designed for VFGNN. Therefore, in our research, we had to adopt some existing backdoor attack methods to evaluate the security of VFGNN. However, these methods are not entirely suitable for VFGNN, so we made appropriate adjustments to them. Below, we provide a detailed description of these adjustments.

\textbf{GF} \cite{chen2022graph} is an adversarial attack method designed for VFGNN. In this method, the attacker first extracts global node embeddings from the system and constructs a shadow model that simulates the active party. Then, noise is added to the node embeddings, which are fed into the shadow model to explore vulnerabilities. Finally, adversarial samples are generated using the gradients derived from the shadow model. However, the objective of adversarial attacks is to misclassify the target node, whereas backdoor attacks aim to classify the target node into a specific class through certain trigger mechanisms. To adapt GF for backdoor attacks, we modified the final step by changing the optimization objective from minimizing the loss of an incorrect class to minimizing the loss of the backdoor class. All hyperparameters for GF remain consistent with those in the original paper.

\textbf{TECB} \cite{chen2023practical} is a backdoor attack method designed for VFL, with the trigger applied to the input samples. When the attacker intends to launch an attack, a trigger is added to the samples fed into the VFL model. However, this method was originally designed for image datasets, and its trigger pattern is not compatible with GNNs, which process graph-structured data. To adapt it to graph data, we modified the approach by setting the attributes of the target node to match the trigger embedding. 
TECB is configured with 4 known target samples, while all other hyperparameter settings remain consistent with the original paper.

\textbf{VILLAIN} \cite{bai2023villain} is a backdoor attack method designed for VFL, where the trigger is applied to the output $H$ of the base model. This method assumes that the attacker must be a participant in the VFL (e.g., an institution) rather than a regular user, as ordinary users cannot modify the intermediate data exchanged between VFL participants. Despite the strong assumption, this design effectively avoids data type mismatches and can be directly applied to VFGNN scenarios. To evaluate its performance, we adjusted the hyperparameters of the original method, setting the trigger magnitude $\beta$ to 4, which is 10 times the default value in the original paper, to further enhance the attack's effectiveness.

\textbf{Edge Triggering (ET)}:
Inspired by previous backdoor attack methods for GNNs \cite{yang2022transferable}, we designed a method to inject backdoors by modifying the graph's structure rather than node features. In this approach, the attacker connects the target node to known nodes of the target class via edges, forming a trigger relationship. This design leverages the GNN's sensitivity to neighboring node features. Consistent with the BVG method, we assume the knowledge of 4 target class nodes. During training, these 4 nodes are interconnected, and during trigger injection, the target node is connected to these 4 nodes to carry out the backdoor attack.

\textbf{Node Feature Triggering (NFT)}:
Complementing the structure-based ET method, we designed a backdoor attack method that focuses solely on node features. Unlike the BVG method, where the trigger is optimized during training, we consider a fixed trigger pattern. A pre-designed trigger is injected into nodes with known labels, and the trigger pattern is consistent with that described in \cite{bai2023villain}. To enhance the attack effect, we set the trigger magnitude $\beta$ to 4, we assume the knowledge of 4 target class nodes.

\textbf{Node Feature Replacement (NFR)}:
We further explore an extreme case where the attacker directly replaces the features of the target node with those of the target class node. This method reflects the sensitivity of VFGNN to node features. The feature replacement operation is expected to cause the target node to be misclassified into the target class, achieving a successful backdoor attack.

\subsection{Performance of Comparison Methods}\label{ComparisonMethods}

For the \textbf{GF} method, as it is fundamentally designed for adversarial tasks, its main task accuracy (MTA) essentially reflects its adversarial attack performance. In adversarial tasks, a lower MTA is desirable, which significantly differs from the goal of backdoor tasks—where maintaining a high MTA without affecting normal task performance is preferred. For this reason, we did not directly compare the MTA of the GF method in the paper, focusing instead on its attack success rate (ASR). From the perspective of ASR, the GF method demonstrates that adversarial attacks can effectively misclassify target nodes. In certain models and datasets, its ASR exceeds 50\%. However, this result still falls short of achieving a practical backdoor attack, primarily because the adversarial performance of the GF method itself is suboptimal.

For the \textbf{TECB} method, we observed that its MTA and ASR are almost mutually exclusive: when the MTA is high, the ASR tends to be low, and vice versa. This characteristic makes TECB highly impractical in VFGNN scenarios. We believe the main reason for this is TECB's second-stage poisoning of the top model. The original method targets image datasets, where the large parameter space and relative robustness of image data make poisoning effective for enhancing backdoor performance. However, in VFGNN, the smaller data and model parameter sizes significantly reduce the stability of the entire model during poisoning, leading to poor performance.

For the \textbf{VILLAIN} method, its backdoor task ASR and main task MTA both perform well, but overall, it is slightly inferior to the BVG method. Additionally, due to the design characteristics of its trigger mechanism, modifications to the intermediate feature \( H \) are highly significant, greatly reducing the stealthiness of the backdoor. As shown in Figure \ref{fig:mse_comparison}, such noticeable feature changes make it challenging to evade detection in certain real-world scenarios.

For the \textbf{ET} method, as only minor adjustments to the graph topology were made during training, its impact on main task performance is minimal. However, the performance of its backdoor task is generally poor, with relatively good attack success only on the Pubmed dataset.

For the \textbf{NFT} method, its MTA performance is similar to that of the ET method. Since the trigger is injected into the feature of only one node at a time, its main task performance degradation is minimal. In terms of backdoor tasks, NFT demonstrates good effectiveness. However, it is worth noting that we set the trigger magnitude \( \beta \) to 4, which is a very large value. We compared the trigger value range of NFT with that of BVG triggers and the input feature range (results are shown in Table \ref{tab:ValueRange}). It can be seen that when \( \beta \) is set to 4 for NFT, its trigger value range almost exceeds the normal range of input features, which could be considered illegal input in real-world scenarios.

For the \textbf{NFR} method, the arbitrary replacement of node features leads to a significant drop in main task performance, while its backdoor task performance is also poor. This indicates that the method struggles to maintain model stability while achieving effective attacks.

\begin{table}
\centering
\begin{tabular}{ccccc}
\toprule
\textbf{Bottom} & \multirow{2}{*}{\textbf{Datasets}} & \textbf{Origin} & \textbf{BVG} & \textbf{NFT} \\
\textbf{Model} &  & \textbf{Data} & \textbf{Trigger} & \textbf{Trigger} \\ \midrule
\multirow{3}{*}{GCN}     & Cora      & 1.0000 & 0.1000 & 0.5916 \\ 
 & Cora\_ml  & 0.3623 & 0.0100 & 0.0982 \\ 
 & Pubmed    & 0.0729 & 0.0100 & 0.2017 \\ \midrule
\multirow{3}{*}{GAT} & Cora      & 1.0000 & 0.0800 & 0.5916 \\ 
 & Cora\_ml  & 0.3623 & 0.0100 & 0.0982 \\ 
 & Pubmed    & 0.0729 & 0.0100 & 0.2017 \\ \midrule
\multirow{3}{*}{SGC} & Cora      & 1.0000 & 0.0800 & 0.5916 \\ 
 & Cora\_ml  & 0.3623 & 0.0100 & 0.0982 \\ 
 & Pubmed    & 0.0729 & 0.0100 & 0.2017 \\ 
\bottomrule
\end{tabular}
\caption{Comparison of the Original Data Value Range with BVG Trigger and NFT Trigger Value Ranges}
\label{tab:ValueRange}
\end{table}

\section{Analysis of Hyperparameter in BVG}

\subsection{Impact of Multi-hop Trigger in BVG} 

\begin{table}[tbp]
\centering
\begin{tabular}{lcccc}
\toprule
\textbf{Bottom} & \multirow{2}{*}{\textbf{Datasets}} & \textbf{ASR} & \textbf{ASR} & \textbf{ASR} \\
\textbf{Model} &  & \textbf{(0 hop)} & \textbf{(1 hop)} & \textbf{(2 hops)} \\
\midrule
\multirow{3}{*}{GCN} & Cora & 74.13 & 98.23 & \textbf{98.86} \\
 & Cora\_ml & 58.30 & 99.80 & \textbf{99.98} \\
 & Pubmed & 80.13 & \textbf{100.00} & \textbf{100.00} \\
\midrule
\multirow{3}{*}{GAT} & Cora & 61.27 & 89.47 & \textbf{99.12} \\
 & Cora\_ml & 41.23 & 86.00 & \textbf{99.92} \\
 & Pubmed & 89.08 & \textbf{100.00} & \textbf{100.00} \\
\midrule
\multirow{3}{*}{SGC} & Cora & 89.08 & 99.14 & \textbf{99.76} \\
 & Cora\_ml & 59.53 & 97.20 & \textbf{99.98} \\
 & Pubmed & 93.32 & 99.50 & \textbf{99.90} \\
\bottomrule
\end{tabular}
\caption{Performance Impact of Multi-Hop Triggers on BVG Method (Best results are highlighted in bold.)}
\label{tab:hop}
\end{table}

Considering the sensitivity of GNN models to spatial relationships, we employed multi-hop triggers in the BVG method. This involves injecting triggers into the multi-hop neighbors of the target node to enhance the performance of backdoor attacks. Since the Bottom model used in our experiment is a 2-layer GNN, its receptive field is limited to 2-hop neighbors. Therefore, we only considered the trigger injection up to 2 hops. To evaluate the impact of multi-hop triggers on BVG performance, we compared the backdoor task performance of the BVG method under three scenarios: without using multi-hop triggers (0 hop), considering only 1-hop neighbors (1 hop), and considering 2-hop neighbors (2 hops). The results are shown in Table \ref{tab:hop}.

We also conducted experiments on three datasets using three bottom models, with each value being the average of five independent trials. It can be observed that the more hops considered for the trigger, the better the ASR performance. Compared to not using multi-hop triggers, the performance improvement when using 2-hop triggers can even exceed 20\%. This demonstrates the effectiveness of multi-hop triggers for VFGNN backdoor attacks.


\subsection{Impact of Backdoor Retention}

\begin{figure}[!tbp]
\centerline{\includegraphics[width=\linewidth]{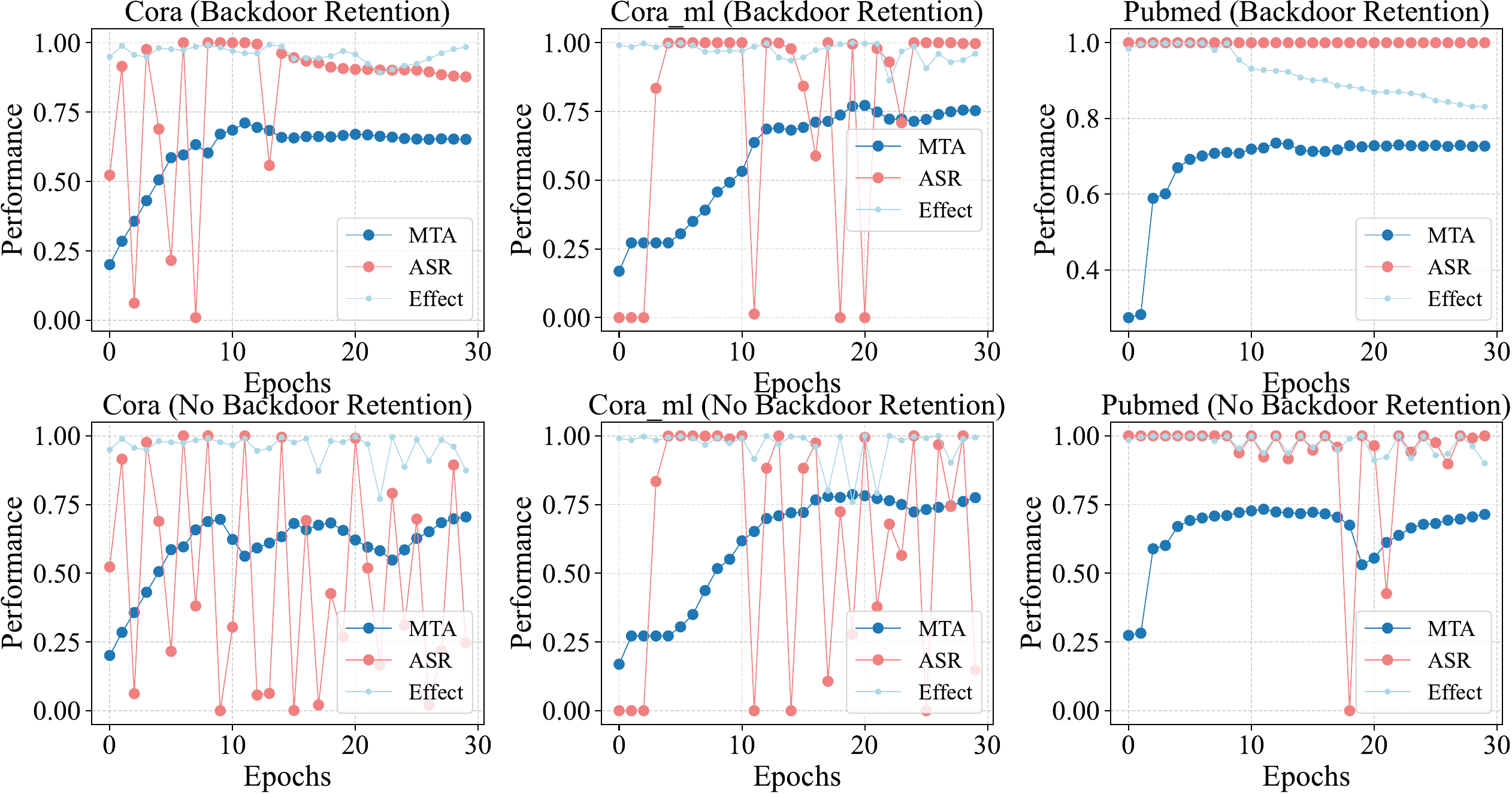}}
\caption{Impact of Backdoor Retention on Attack Stability Across Different Datasets. (GCN)}
\label{fig:line_gcn}
\end{figure}

\begin{figure}[!tbp]
\centerline{\includegraphics[width=\linewidth]{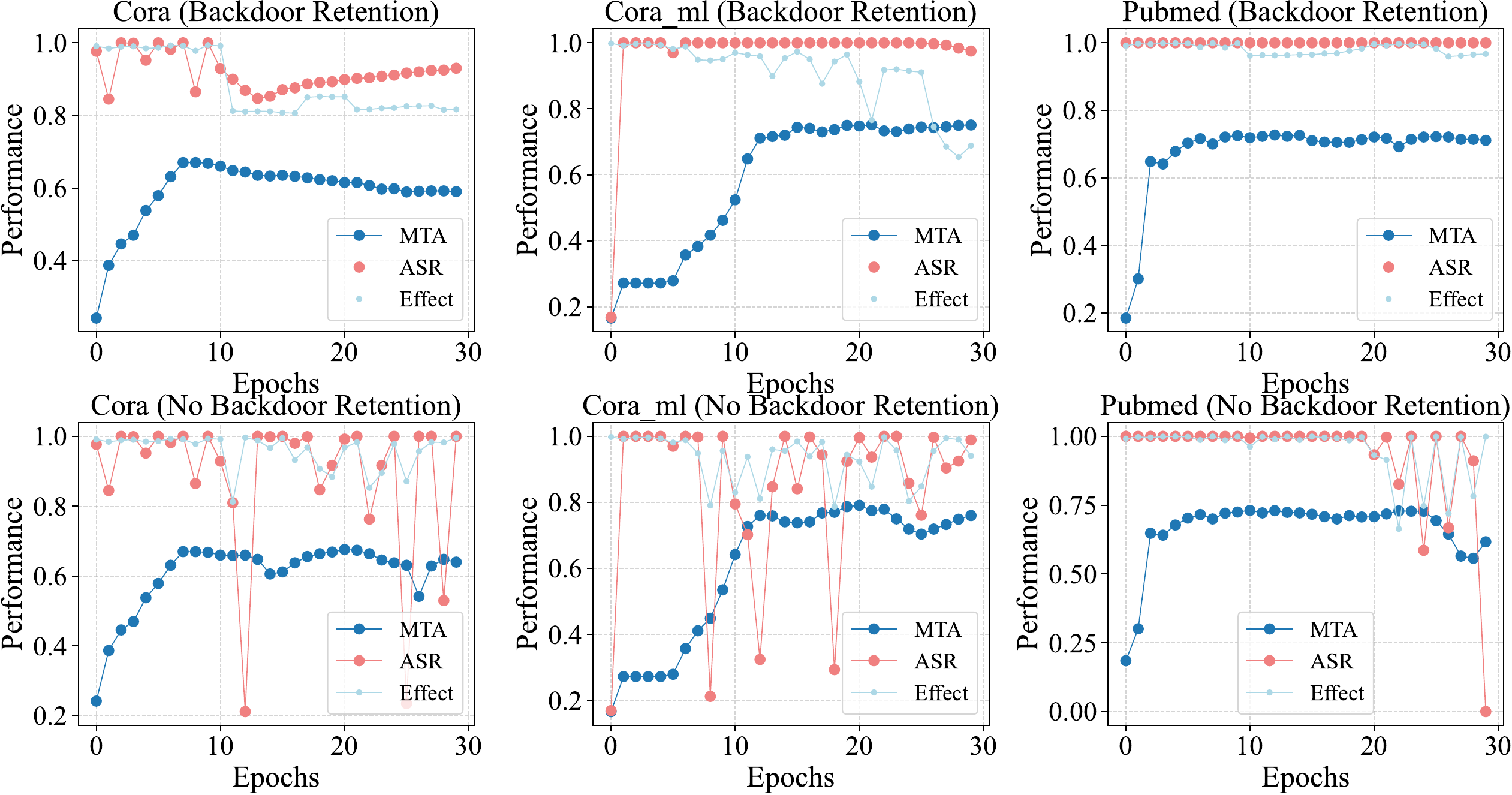}}
\caption{Impact of Backdoor Retention on Attack Stability Across Different Datasets. (GAT)}
\label{fig:line_gat}
\end{figure}

\begin{figure}[!tbp]
\centerline{\includegraphics[width=\linewidth]{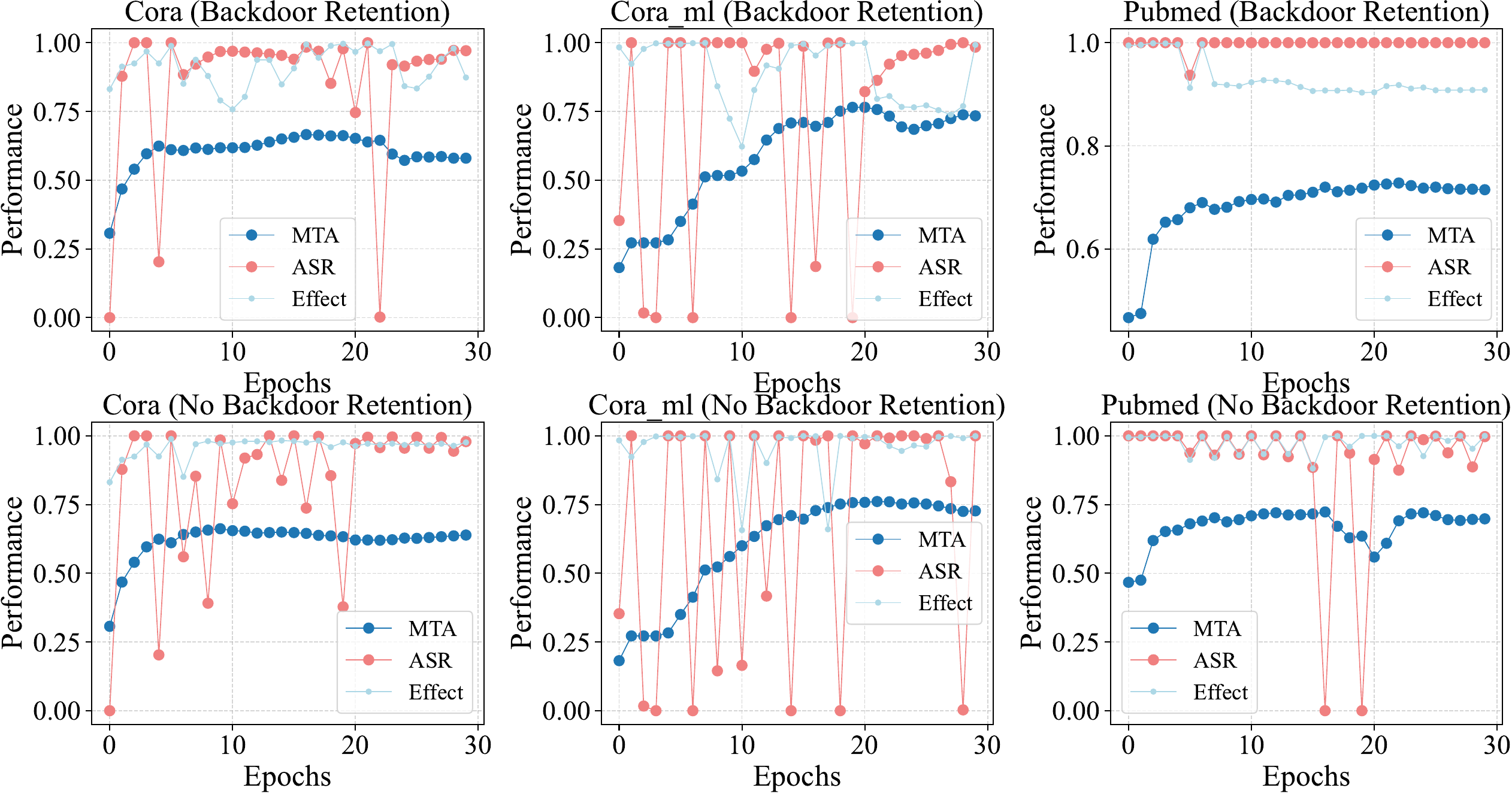}}
\caption{Impact of Backdoor Retention on Attack Stability Across Different Datasets. (SGC)}
\label{fig:line_sgc}
\end{figure}

To visually demonstrate the effect of Backdoor Retention, we conducted a total of nine experiments across three bottom models and three datasets, comparing the results with and without Backdoor Retention. In each experiment, we recorded the main task MTA, backdoor task ASR, and the current backdoor effect (effect) measured by the attacker after each epoch. The experimental results with GCN as the bottom model are shown in Figure \ref{fig:line_gcn}, GAT results in Figure \ref{fig:line_gat}, and SGC results in Figure \ref{fig:line_sgc}.

From the experiments without Backdoor Retention, we observed significant fluctuations in the backdoor task performance. This is mainly because the number of samples with known labels is too small, and their proportion is much lower compared to the thousands or tens of thousands of samples, making the trigger signal prone to being "forgotten." The evaluation of backdoor effect (effect) shows a strong correlation with the backdoor task performance, with its fluctuations almost perfectly reflecting the changes in the backdoor task.

In the experiments with Backdoor Retention, we observed a significant improvement in the stability of the backdoor task. This is particularly evident when GAT is used as the bottom model, where the backdoor performance fluctuations were almost completely eliminated. For GCN and SGC, the stability of the backdoor task also showed a clear improvement. This effect is primarily achieved because the attacker, through evaluating the backdoor effect, discards updates that negatively impact the backdoor effect and only accepts updates beneficial to the backdoor task, thereby ensuring the stability of the backdoor.

\subsection{Impact of Target Dataset Size} 

\begin{figure}[!tbp]
\centerline{\includegraphics[width=1\linewidth]{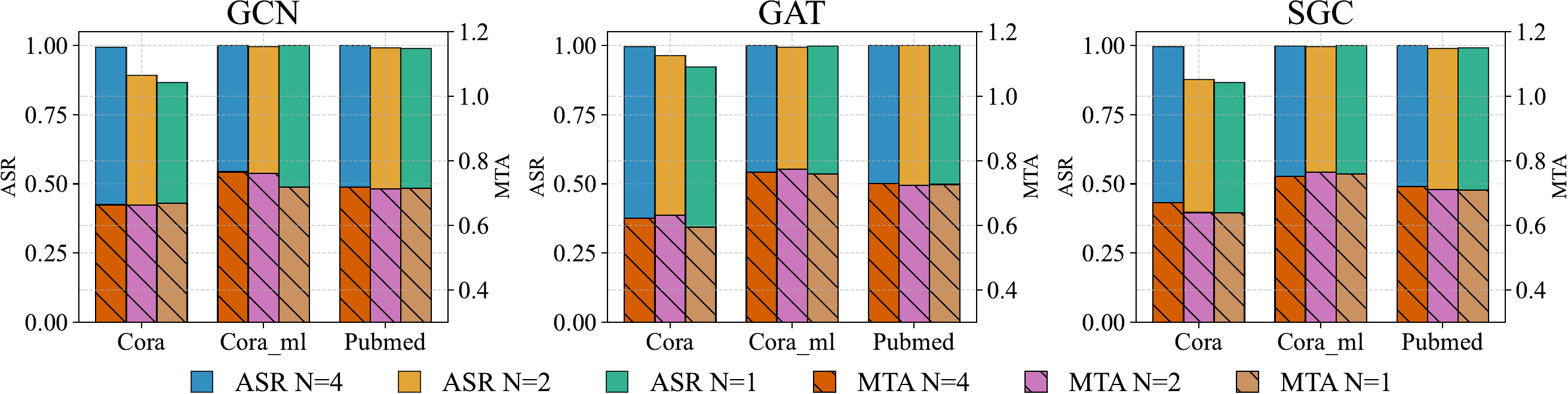}}
\caption{Impact of Different Sizes of $\mathcal{V}_p$ on BVG Performance (ASR and MTA) Across Three Datasets.}
\label{fig:nums}
\end{figure}

The BVG method requires a minimal amount of auxiliary target sample labels for backdoor attacks. These labeled nodes play a crucial role in BVG, making the size of $\mathcal{V}_p$ (denoted as $N$) an important hyperparameter that affects BVG's effectiveness and its real-world threat. We measured the impact of different sizes of $\mathcal{V}_p$ on BVG performance (ASR and MTA) across three base models and three datasets, with the results plotted in Figure \ref{fig:nums}.

The figure shows that the BVG method can achieve excellent backdoor attack performance even with very limited auxiliary target data. In the Cora\_ml and Pubmed datasets, the BVG method performs well with just one target class node. For more complex datasets like Cora, the method achieves over 80\% attack accuracy with just one target class node, and only four target class nodes are needed to achieve good ASR and MTA performance. This demonstrates that BVG can effectively carry out backdoor attacks in VFGNN, even with very limited information about target samples, thus providing an efficient attack method in practical VFGNN scenarios.

\subsection{Impact of Trigger Perturbation} 

\begin{figure}[]
\centerline{\includegraphics[width=1\linewidth]{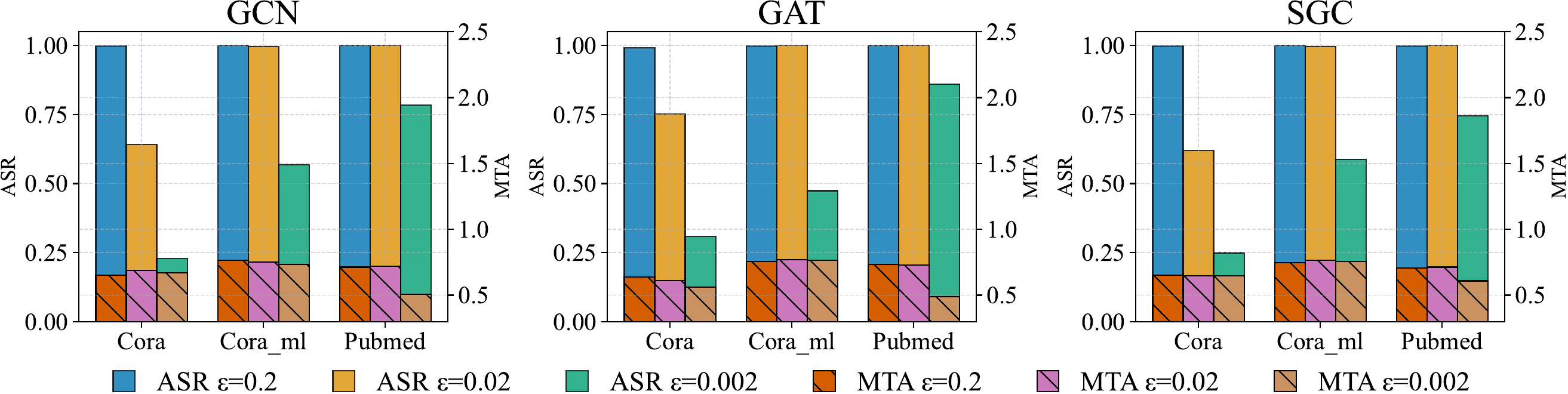}}
\caption{Performance of BVG on Three Datasets and Three Bottom Models with Different Perturbation Levels ($\epsilon$)}
\label{fig:e}
\end{figure}

In backdoor attacks, trigger perturbations are usually constrained to ensure stealthiness. Figure \ref{fig:e} shows the performance of BVG on three datasets and three Bottom models with perturbations ($\epsilon$) of 0.002, 0.02, and 0.2.

We conducted experiments on three datasets and observed that as the perturbation increases, the ASR performance of the backdoor attack significantly improves. The required perturbation value varies across different datasets, with the Cora dataset requiring a larger perturbation than the other datasets. Therefore, within the constraints of ensuring stealthiness, selecting an appropriate range of perturbation values can achieve a higher success rate for backdoor attacks.

\section{Attack Efficacy under Defense}\label{appendix:defen}

\begin{figure}[h]
\centerline{\includegraphics[width=\linewidth]{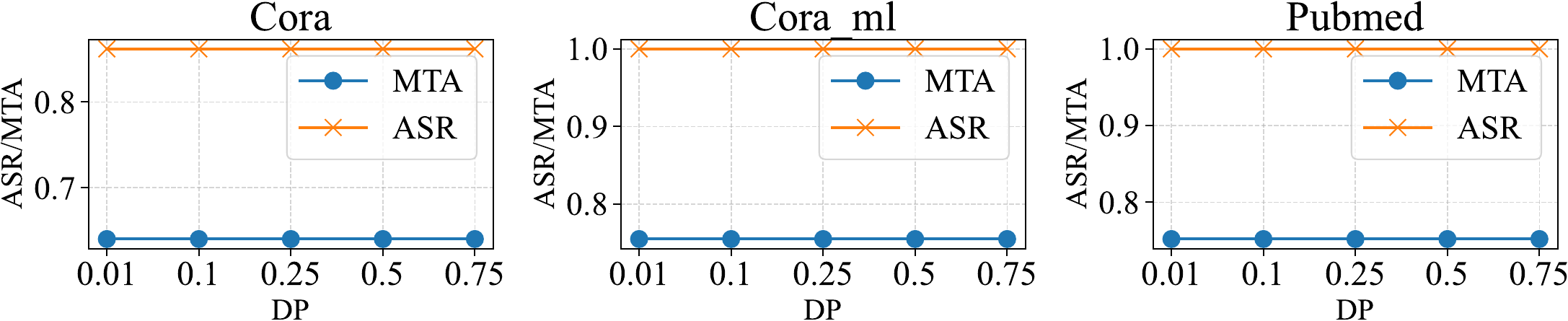}}
\caption{The performance of BVG under DP-SGD defense (GAT)}
\label{fig:DP_gat}
\end{figure}

\begin{figure}[h]
\centerline{\includegraphics[width=\linewidth]{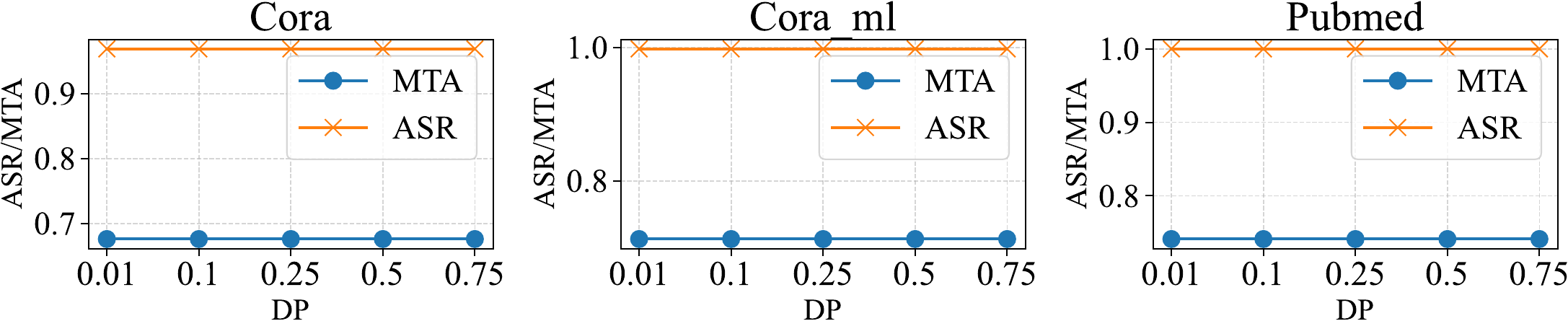}}
\caption{The performance of BVG under DP-SGD defense (SGC)}
\label{fig:DP_sgc}
\end{figure}

\begin{figure}[h]
\centerline{\includegraphics[width=\linewidth]{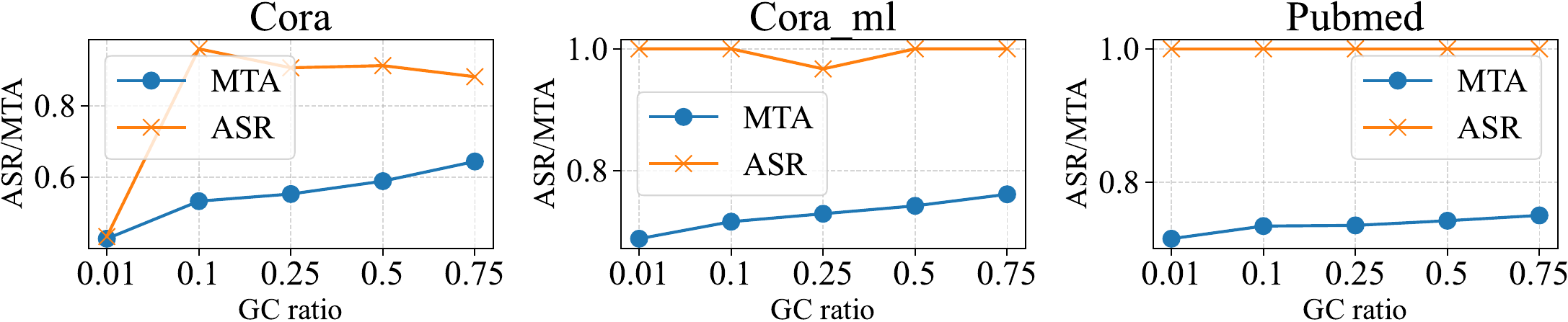}}
\caption{The performance of BVG under GC defense (GAT)}
\label{fig:GC_gat}
\end{figure}

\begin{figure}[h]
\centerline{\includegraphics[width=\linewidth]{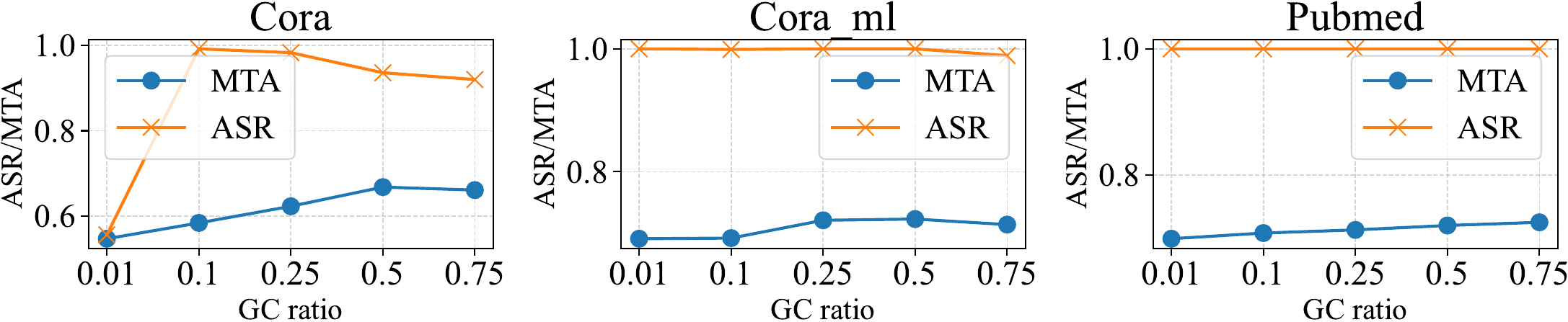}}
\caption{The performance of BVG under GC defense (SGC)}
\label{fig:GC_sgc}
\end{figure}

\begin{figure}[h]
\centerline{\includegraphics[width=\linewidth]{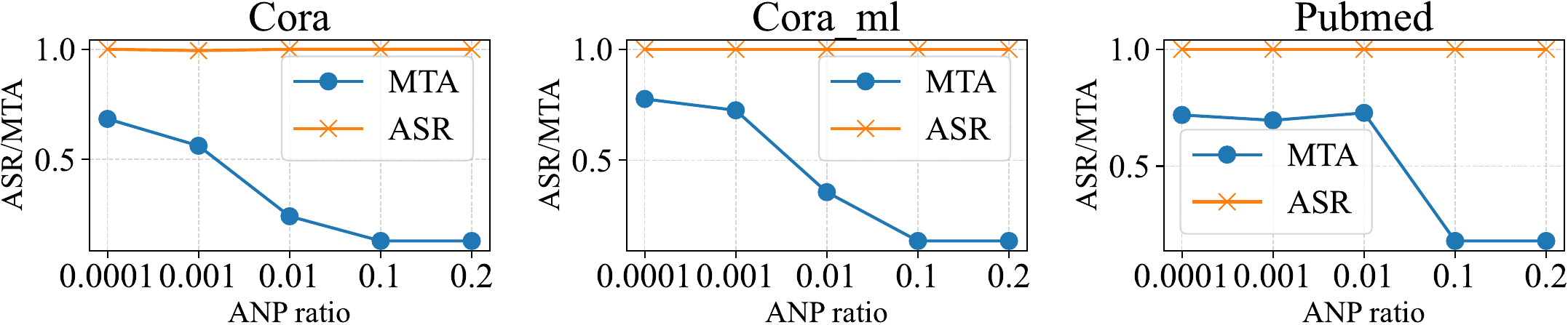}}
\caption{The performance of BVG under ANP defense (GAT)}
\label{fig:ANP_gat}
\end{figure}

\begin{figure}[h]
\centerline{\includegraphics[width=\linewidth]{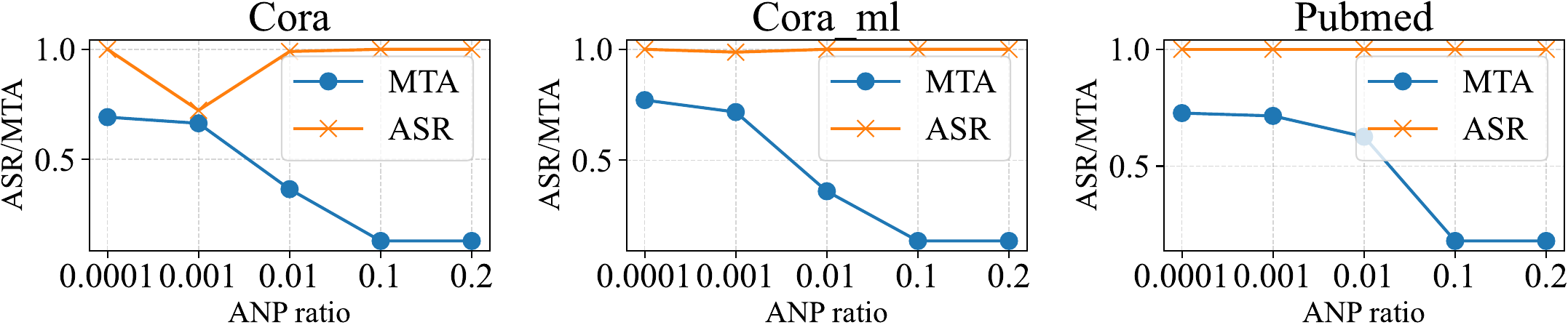}}
\caption{The performance of BVG under ANP defense (SGC)}
\label{fig:ANP_sgc}
\end{figure}

\begin{figure}[h]
\centerline{\includegraphics[width=\linewidth]{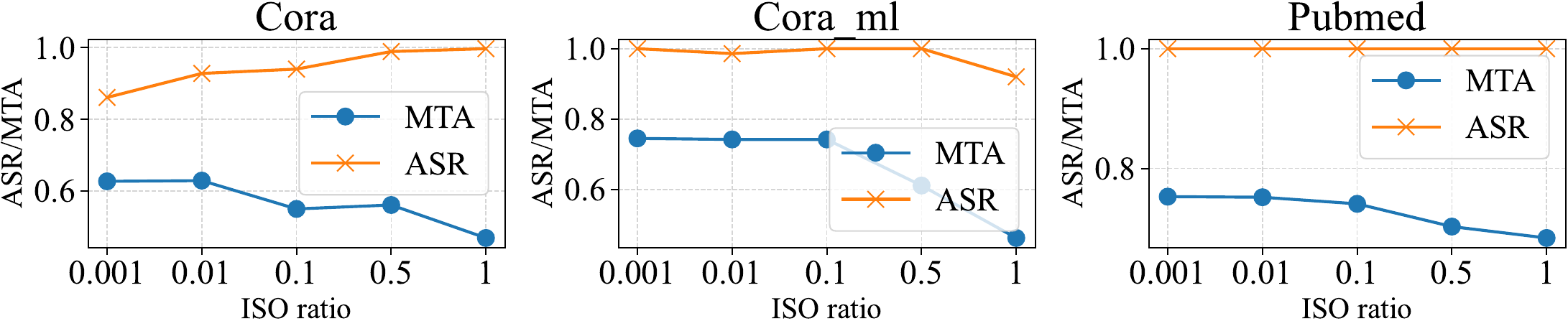}}
\caption{The performance of BVG under ISO defense (GAT)}
\label{fig:ISO_gat}
\end{figure}

\begin{figure}[h]
\centerline{\includegraphics[width=\linewidth]{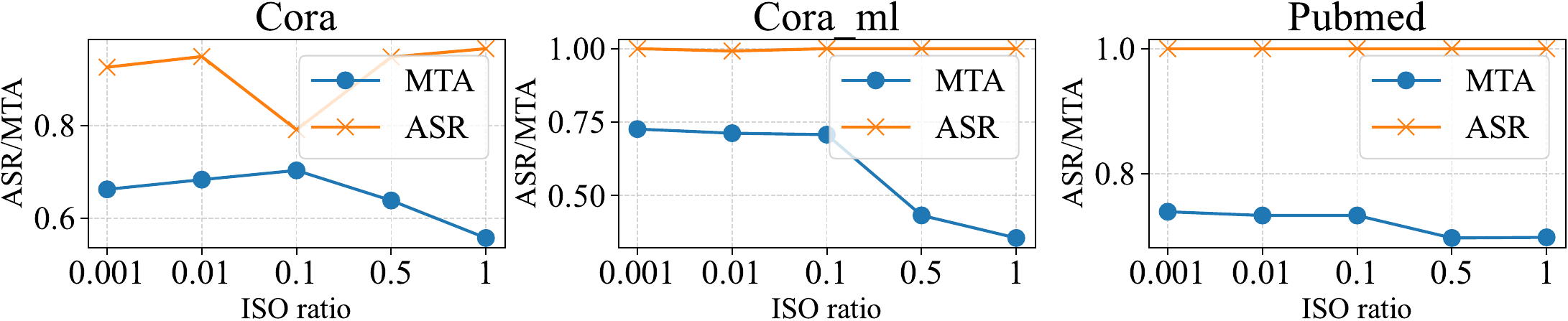}}
\caption{The performance of BVG under ISO defense (SGC)}
\label{fig:ISO_sgc}
\end{figure}

Due to space constraints, we present the backdoor defense results for GAT and SGC as bottom models. Figures \ref{fig:DP_gat} and \ref{fig:DP_sgc} show the defense results of the DP-SGD method, Figures \ref{fig:GC_gat} and \ref{fig:GC_sgc} present the results of the GC method, Figures \ref{fig:ANP_gat} and \ref{fig:ANP_sgc} illustrate the defense performance of the ANP method, and Figures \ref{fig:ISO_gat} and \ref{fig:ISO_sgc} display the results of the ISO method.

The results indicate that different datasets and models respond similarly to these defense methods. DP-SGD and GC, which target label inference attacks, are almost ineffective against our attack. For the backdoor defense methods ANP and ISO, while the backdoor attack success rate decreases as model accuracy drops, these methods are still insufficient to effectively counter our attack.

\end{document}